# Complete-SVD: A Functional Norm Approach to Low-Rank LLM Compression

**Alper Kalle**[†,*] **Theo Rudkiewicz**[‡,*]
**Mohamed-Oumar Ouerfelli**[†] **Mohamed Tamaazousti**[†]
{alper.kalle,mohamed-oumar.ouerfelli,mohamed.tamaazousti}@cea.fr
theo.rudkiewicz@inria.fr

## Abstract

Post-training low-rank compression of neural networks must not only approximate the original weights but also preserve the model's input-output behavior. Recent output-aware methods measure the compression error beyond the compressed layer, at the model output, weighting logit changes by the observed next token or by the model's predicted probabilities. We propose Complete-SVD, a method for large language models that minimizes the $L_2$ distance between the logits of the original and compressed models, weighting all logit changes equally. Linearizing the network around the original weights, together with a Kronecker-factored approximation, reduces this objective to a weighted SVD with a closed-form solution. Its output-side factor is intractable to compute; we estimate it without bias from random probes, at a cost of K backward passes per sequence, independent of vocabulary size and sequence length. At matched settings with uniform ranks and WikiText-2 calibration, Complete-SVD achieves the lowest C4 and PTB perplexity in every setting, and at ratio 0.7 under LEMS rank allocation. Replacing the $L_2$ distance by the Kullback–Leibler divergence (KL-SVD), with the same estimation framework, lowers calibration perplexity in most settings and raises downstream accuracy, but raises cross-corpus perplexity. Complete-SVD also improves IFEval instruction-following accuracy over the strongest prior output-aware methods throughout.

## 1 Introduction and Related Work

Large language models have driven rapid progress across reasoning and generation tasks (Kaplan et al., 2020), but their growing parameter counts make inference prohibitively expensive. Model compression reduces this cost through quantization (Frantar et al., 2023; Lin et al., 2024; Xiao et al., 2023), pruning (Frantar and Alistarh, 2023; Ma et al., 2023; Sun et al., 2024), knowledge distillation (Hinton et al., 2015), or low-rank factorization, which replaces each weight matrix by the product of two smaller matrices. Low-rank factorization retains dense operations that run efficiently on standard hardware. Post-training low-rank methods differ in their per-layer objective, i.e., the criterion minimized by the low-rank approximation of each layer; rank allocation, storage schemes (Wang et al., 2025b), and cross-layer basis sharing (Wang et al., 2025a) are complementary to it. This work concerns the objective, that is, which change in the model the approximation of a layer should penalize.

**Error at the compressed layer.** Let $W$ be the weight matrix of a linear layer and $\widehat{W}$ its approximation of rank at most $r$. Truncated singular value decomposition (SVD) returns the $\widehat{W}$ closest to $W$ in Frobenius norm, which treats all weight directions equally, and replacing the weights of an LLM by these approximations severely degrades its performance. Activation-aware methods instead minimize the error on the layer outputs, $\left\|X\left(W-\widehat{W}\right)\right\|_F$, where the rows of $X$ are input activations collected on a small calibration set of text sequences. ASVD (Yuan et al., 2023) rescales $W$ by per-channel activation magnitudes, and SVD-LLM (Wang et al., 2025c) solves this problem in closed form by whitening with $\left(X^\top X\right)^{\frac{1}{2}}$ before LoRA fine-tuning. For tensor decompositions of convolutional networks, Kalle et al. (2025) cast this principle as a functional norm: the $L_2$ distance, over the data distribution, between the original and compressed networks up to the compressed layer. This norm involves only the upstream network, *i.e.*, the preceding layers, and reduces to the activation-aware objective for a linear layer. The error at the layer output is not, however, the error of the model: it propagates through the downstream network, *i.e.*, the following layers, which amplifies it along some directions and attenuates it along others, and, in a causal transformer, carries it to the predictions at all subsequent positions.

---

[*] Equal contribution.
[†] Université Paris-Saclay, CEA, List, F-91120 Palaiseau, France
[‡] TAU team, LISN, Université Paris-Saclay, CNRS, Inria, 91405 Orsay, France

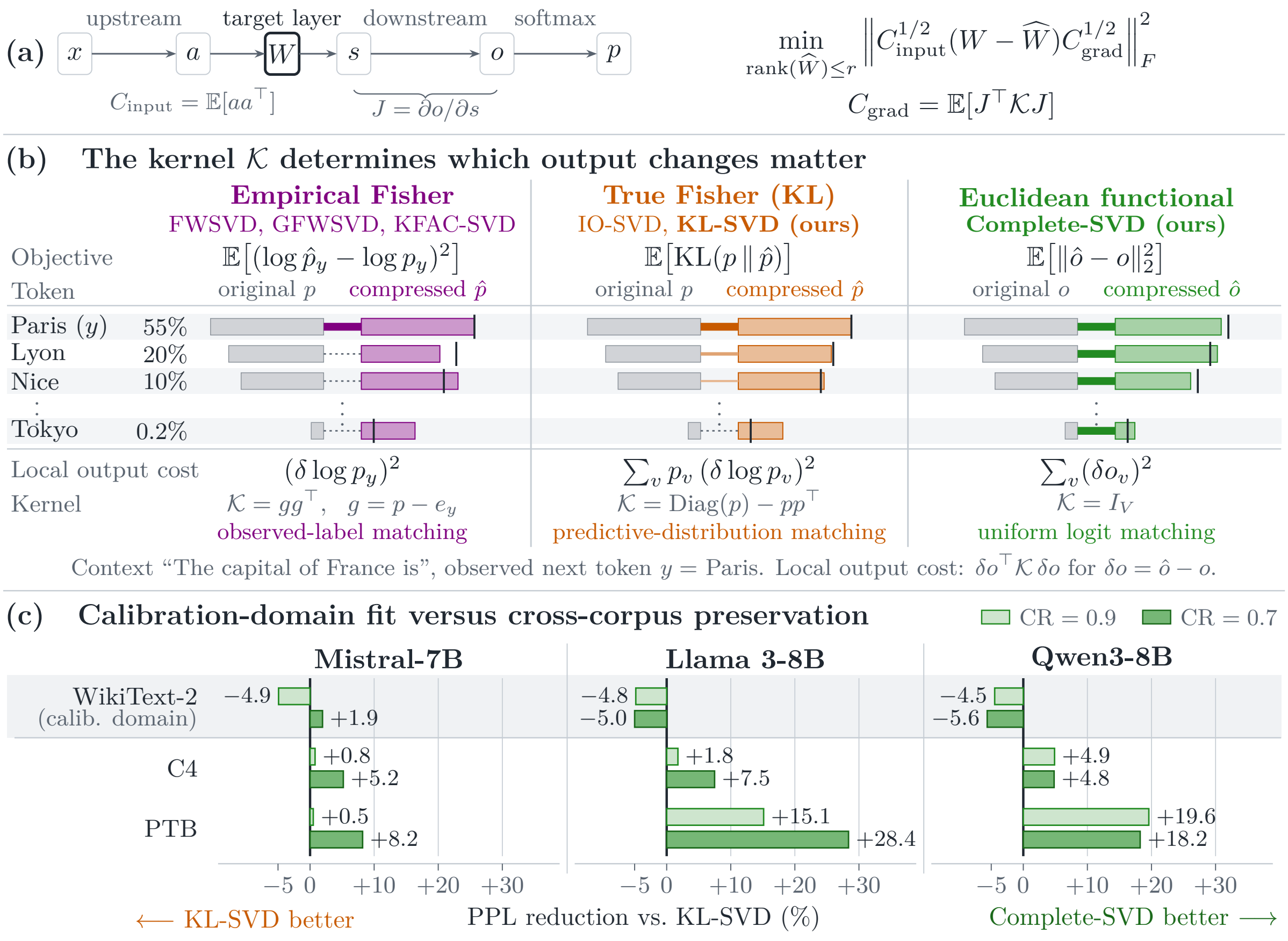

Figure 1: **Output geometry changes the compression tradeoff. (a)** The output-aware methods can be written in this Kronecker-factored form and differ mainly in the output kernel $\mathcal{K}$: $C_{\text{input}} = \Sigma_{\text{input}}$, and $C_{\text{grad}} = \Sigma_{\text{grad}}$ for **Complete-SVD**. **(b)** Illustrative next-token example. Fisher-based objectives compare $p = \mathrm{softmax}(o)$, whereas **Complete-SVD** compares the logits $o$; local output costs hold to leading order in $\delta o$. Link width is the weight of each token in this cost, and black ticks mark original values. Bars share a logarithmic scale ($o = \log p + \text{const}$); compressed values are schematic. **(c)** Perplexity reduction $100(\mathrm{PPL}_{\mathrm{KL}} - \mathrm{PPL}_{\mathrm{Complete}})/\mathrm{PPL}_{\mathrm{KL}}$ of **Complete-SVD** relative to **KL-SVD** at compression ratios (CR) 0.9 and 0.7, with uniform ranks. WikiText-2 is the calibration domain.

**Error at the model output.** Output-aware methods measure the error at the model output. Each compares the outputs of the original and compressed models through a discrepancy, whose local quadratic approximation weights a change of the layer output by its effect on the logits and by the importance that the discrepancy assigns to each logit change (Figure 1, panel b). A first family preserves the training loss through the empirical Fisher matrix, built from loss gradients at the observed next token. FWSVD (Hsu et al., 2022) uses a diagonal approximation, GFWSVD (Chekalina et al., 2026) a Kronecker factorization fitted from a limited number of sequence-level gradients, and KFAC-SVD (Thoma et al., 2026) Kronecker factors estimated from token-level statistics, which alleviate the rank deficiency of GFWSVD. For each token, the empirical Fisher only penalizes the change in the log-probability of the observed next token, a single direction among all possible logit changes (Kunstner et al., 2019). A second family preserves the predictive distribution through the Kullback–Leibler (KL) divergence, whose local quadratic approximation is governed by the true Fisher information (Amari, 1998). IO-SVD (Abbasi et al., 2026) compresses LLMs with this objective. IO-SVD approximates the true Fisher matrix through two reductions: it retains only the effect of a perturbation at a given position on the prediction at that same position, and it restricts the predictive distribution to its most probable tokens, renormalized. Both reductions are deterministic, so the resulting bias does not vanish as the calibration set grows (Appendix D.5.1).

**Complete-SVD.** Our objective is the $L_2$ functional distance between the original and compressed networks, *i.e.*, the mean squared Euclidean distance between their logits over the data distribution. Every logit change is thus penalized with equal weight, whereas the Fisher objectives weight each logit change by the observed next token or by the probabilities that the model predicts on the calibration contexts (Figure 1, panel b). We find that the uniform weighting better preserves perplexity on corpora other than the calibration one (Figure 1, panel c). A Kronecker-factored approximation (K-FAC) (Martens and Grosse, 2015; Eschenhagen et al., 2023) of the objective reduces the objective to a weighted SVD with a closed-form solution. Its two weights are the input covariance $\Sigma_{\text{input}}$ of activation-aware methods and the Gram matrix $\Sigma_{\text{grad}}$ of the logit Jacobians, which measures how strongly each direction of the layer output affects the logits.

**Tractable estimation.** The choice of objective also determines the cost of the output-side weight. The empirical Fisher is inexpensive because it involves only the loss, a single scalar whose gradient takes one backward pass per sequence. The KL and $L_2$ objectives involve every logit at every position, and computing $\Sigma_{\text{grad}}$ exactly takes one backward pass per vocabulary entry and per position, tens of millions per sequence for Llama 2-7B. The linearity of the LM head absorbs the vocabulary exactly, without any estimation, into the $D \times D$ Gram matrix of the LM-head weights, with $D$ the hidden dimension: the number of backward passes per output position drops from $V$ to $D$, a factor of about 31 for Llama 3-8B ($V \approx 128\text{k}$, $D = 4096$), and the resulting expression of $\Sigma_{\text{grad}}$ is still exact. Random probes with random signs on the output positions then yield an unbiased estimator of $\Sigma_{\text{grad}}$ costing $K$ backward passes per sequence, independently of the vocabulary size and the sequence length. With probes drawn from the predictive distribution, the same construction gives an unbiased estimator of the true Fisher, which yields **KL-SVD**, the KL objective of IO-SVD without its reductions (see Appendix D.5.1 for details). Comparing **KL-SVD** with **Complete-SVD** then separates the choice of objective from the approximation used to compute it.

Our contributions are the following.

- **Complete functional norm.** We extend the functional-norm objective of Kalle et al. (2025) from the upstream network to the full model, and reduce it to a closed-form weighted SVD that penalizes all logit changes equally (Section 2.2).
- **An unbiased estimator of $\Sigma_{\text{grad}}$** whose cost is $K$ backward passes per sequence, independently of the vocabulary size and the sequence length, validated against the exact matrix at reduced scale (Section 2.3). Drawing the probes from the predictive distribution, the same construction yields an unbiased estimator of the true Fisher matrix $F_s$ at the same cost. Substituting it for $\Sigma_{\text{grad}}$ gives **KL-SVD**, the KL objective of IO-SVD without its reductions, which we show to be deterministic deviations from $F_s$ that persist for any calibration budget (Appendix D.5). **KL-SVD** shares every other component with **Complete-SVD** and therefore isolates the effect of the objective from that of its estimation.
- **Empirical evaluation.** On Mistral-7B, Llama 3-8B and Qwen3-8B calibrated on WikiText-2 with uniform ranks, **Complete-SVD** achieves the lowest C4 and PTB perplexity among all compared methods at both tested compression ratios (0.9 and 0.7). On Llama 3-8B at ratio 0.9, it reduces PTB perplexity from 30.9 with IO-SVD, the strongest prior baseline, to 23.0. Without fine-tuning and combined with GPTQ, **Complete-SVD** compresses Llama 2-7B from 13.5 GB to 1.9 GB at a WikiText-2 perplexity of 16.7 (Section 3).

## 2 Method

### 2.1 Problem Statement

We suppose that we are in a supervised learning setting, where we have a data distribution $\mathcal{D}$ on $\mathcal{X} \times \mathcal{Y}$ and a per-sample loss function $\ell : \mathcal{Y} \times \mathcal{Y} \to \mathbb{R}_+$ (for classification this is typically the cross-entropy loss and $\mathcal{Y} = \mathbb{R}^V$ with $V$ the number of classes). The goal of supervised learning is to find a function $f_W : \mathcal{X} \to \mathcal{Y}$ such that the expected loss $\mathcal{L}(f_W) = \mathbb{E}_{x \sim \mathcal{D}}(\ell(f_W(x), y))$ is small. We suppose that we already have such a function $f_W$ that has low expected loss. We can decompose this function as a pre-computation $a$, a fully-connected layer with weights $W$ ($\mathrm{L}_W$) and a post-computation $q$ such that $f_W = q \circ \mathrm{L}_W \circ a$:

$$f_W : x \overset{a}{\mapsto} a(x) \overset{\mathrm{L}_W}{\longmapsto} s(x) \overset{q}{\mapsto} o \qquad (1)$$

with $x \in \mathcal{X},\ a(x) \in \mathbb{R}^{n_a},\ s(x) \in \mathbb{R}^{n_s},\ o \in \mathbb{R}^V$. We aim to replace the fully-connected layer with two fully-connected layers $\mathrm{L}_{W_1}$ and $\mathrm{L}_{W_2}$ such that $f_W \approx q \circ \mathrm{L}_{W_2} \circ \mathrm{L}_{W_1} \circ a$ with the constraint that the intermediate dimension is smaller than a specific value $k$ and the expected loss of this new function is close to the expected loss of $f_W$. We note $\widehat{W} = W_1 \times W_2$ the effective weight of the new function. In practice, we aim to solve:

$$\operatorname*{argmin}_{\widehat{W}} \mathcal{L}(f_{\widehat{W}}) \text{ such that } \operatorname{rank}(\widehat{W}) \leq k. \quad (2)$$

### 2.2 Norm Optimization

We minimize the $L_2$ functional distance $\|f_W - f_{\widehat{W}}\|^2$; alternative objectives (KL, empirical Fisher) are derived in Appendix B. Since optimizing this distance directly is intractable, we linearize $f_{\widehat{W}}$ around $W$.

**Theorem 1.** *With the notation introduced above, let* $\Sigma_{\text{input}} := \mathbb{E}_{x\sim\mathcal{D}}[a(x)(a(x))^\top]$ *denote the input covariance and* $\Sigma_{\text{grad}} := \mathbb{E}_{x\sim\mathcal{D}}\left[\left(\frac{\partial f_W}{\partial s}(x)\right)^\top \frac{\partial f_W}{\partial s}(x)\right]$ *the Gram matrix of the output Jacobians at the compressed layer. Then:*

(i) Exact first-order expansion. *Let* $G_W := \mathbb{E}_{x\sim\mathcal{D}}\left[\left(\frac{\partial f_W(x)}{\partial W}\right)^\top \frac{\partial f_W(x)}{\partial W}\right] \in \mathbb{R}^{n_s n_a \times n_s n_a}$. *Then*

$$\|f_W - f_{\widehat{W}}\| = \left\|G_W^{\frac{1}{2}} \overrightarrow{\mathrm{d}W}\right\|_F + o(\|\mathrm{d}W\|_F). \quad (3)$$

(ii) Kronecker-factored approximation. *Under the K-FAC structural approximation* $G_W \approx \Sigma_{\text{input}} \otimes \Sigma_{\text{grad}}$*, the right-hand side of Equation 3 admits the closed-form approximation*

$$\left\|G_W^{\frac{1}{2}} \overrightarrow{\mathrm{d}W}\right\|_F \approx \left\|\Sigma_{\text{input}}^{\frac{1}{2}} \mathrm{d}W \Sigma_{\text{grad}}^{\frac{1}{2}}\right\|_F. \quad (4)$$

**Proof Sketch.** Linearizing $f_W$ around $W$ gives $\|f_W - f_{\widehat{W}}\| \approx \left\|G_W^{\frac{1}{2}} \overrightarrow{\mathrm{d}W}\right\|_F$ with $G_W = \frac{\partial f_W}{\partial W}^T \frac{\partial f_W}{\partial W}$. Approximating the Gram matrix $G_W \approx \Sigma_{\text{input}} \otimes \Sigma_{\text{grad}}$ via K-FAC (Dangel et al., 2025) and leveraging the Kronecker property $(X \otimes Y)\ \mathrm{vec}(Z) = \mathrm{vec}(YZX^T)$ yields the weighted Frobenius norm. See Appendix A for the full derivation. This optimization problem is solvable efficiently in closed-form via SVD. See Appendix F for the computation of $\Sigma_{\text{input}}^{\frac{1}{2}}$ and $\Sigma_{\text{grad}}^{\frac{1}{2}}$.

### 2.3 Tractable Computation of $\Sigma_{\text{grad}}$

Theorem 1 reduces compression to the closed-form weighted Frobenius norm, but the computation of output-Jacobian Gram matrix $\Sigma_{\text{grad}}$ requires one backward pass per output dimension, i.e., $V$ backward passes per token which is intractable for $V \approx$ 32k–128k in modern LLMs.

**LM-head factorization.** Let $f = W_U h$ be the predicted logits with $W_U \in \mathbb{R}^{V\times D}$ the LM-head weight and $h$ the last hidden state. Using the chain rule, $\Sigma_{\text{grad}}$ is a sum of $\left(\partial h_{n,t'}/\partial s_{n,t}\right)^\top W_U^\top W_U \left(\partial h_{n,t'}/\partial s_{n,t}\right)$ terms, where $\partial h_{n,t'}/\partial s_{n,t}$ is the sensitivity of the hidden state at output position $t'$ to position $t$. The vocabulary dimension is absorbed once into the precomputed Gram matrix $M_G := W_U^\top W_U$, reducing the number of cotangent directions from $V$ to $D$, a factor $\approx 8$ for Llama 2-7B and $\approx 31$ for Llama 3-8B.

**Broadcast backward.** The remaining factor $D$ is the hidden size, and still dominates the calibration budget. Drawing a Rademacher matrix $\Psi \in \{\pm 1\}^{V\times K}$ and setting $Z := W_U^\top \Psi$ gives $K \leq D$ cotangent directions $z_k$ with $\mathbb{E}_\Psi\left[z_k z_k^\top\right] = M_G$. Each direction is broadcast to every position of the LM-head input, weighted by an independent random sign $\varepsilon_{n,t'}^{(k)}$, and propagated in a single backward pass: the broadcast accumulates the contribution of all positions in one traversal of the computation graph, while the signs annihilate the cross-position terms in expectation.

Together these give the broadcast estimator $\tilde{\Sigma}_{\text{grad}}$, costing one forward pass and $K$ backward passes per sequence, independent of both $V$ and $T_n$, and unbiased for $\Sigma_{\text{grad}}$ conditionally on the calibration set (Proposition 4). Against the exact $\Sigma_{\text{grad}}$, computable at reduced scale, $\tilde{\Sigma}_{\text{grad}}$ recovers most of the compression quality (Table 13).

**KL-SVD.** With probes drawn from the predictive distribution instead of Rademacher columns, the same construction yields an unbiased estimator, at the same cost, of the true Fisher matrix $F_s = \mathbb{E}[J^\top(\mathrm{diag}(p) - pp^\top)J]$, where $J := \frac{\partial f_W}{\partial s}$, of the KL divergence between the predictive distributions (Appendix D.5). Substituting it for $\Sigma_{\text{grad}}$ yields **KL-SVD**, all other components are shared with **Complete-SVD** (Figure 1).

## 3 Experiments

We compare our method, **Complete-SVD**, against state-of-the-art SVD-based compression methods, including standard SVD, ASVD (Yuan et al., 2023), FWSVD (Hsu et al., 2022), GFWSVD (Chekalina et al., 2026), SVD-LLM (Wang et al., 2025c), Dobi-SVD (Wang et al., 2025b), KFAC-SVD (Thoma et al., 2026), and IO-SVD (Abbasi et al., 2026) as well as structured pruning methods such as LLM-Pruner (Ma et al., 2023) and Wanda-SP (Sun et al., 2024). We further demonstrate that **Complete-SVD** is compatible with state-of-the-art quantization methods by combining it with GPTQ (Frantar et al., 2023). To validate our approach, we conduct experiments on Llama 2-7B (Touvron et al., 2023) Base and Chat, Llama 3-8B (Grattafiori et al., 2024), Mistral-7B (Jiang et al., 2023) and Qwen3-8B (Yang et al., 2025), evaluating on the language modeling datasets WikiText-2 (Merity et al., 2017), C4 (Raffel et al., 2020), and PTB (Marcus et al., 1993), the downstream benchmarks OpenbookQA (Mihaylov et al., 2018), WinoGrande (Sakaguchi et al., 2021), HellaSwag (Zellers et al., 2019), PIQA (Bisk et al., 2020), ARC-c, and ARC-e (Clark et al., 2018), and the generation dataset IFEval (Zhou et al., 2023). The covariance matrices $\Sigma_{\text{grad}}$ and $\Sigma_{\text{input}}$ are computed from 256 randomly sampled sequences of length 2048 from WikiText-2 dataset (See Appendix D.3 for the calibration cost). Each experiment runs on a single GPU (NVIDIA H100 or H200), except the inference-efficiency measurements (Appendix G). The compression ratio (CR) is the parameter count of the compressed layers divided by that of the original layers.

**LEMS Rank allocation.** To test whether **Complete-SVD** composes with a published rank allocator, we use LEMS (Thoma et al., 2026) with the authors' procedure: one KL probe per layer, with that layer reduced to 10% of its parameters, on 32 calibration windows of WikiText-2; the harmonic cross-layer term; ranks in multiples of 8; and an exact integer-program solve (HiGHS) (Huangfu and Hall, 2018) at the target budget. Ranks are allocated under each method's own whitening, so every method receives the allocation LEMS finds for it, and all use KFAC-SVD's damping. We fix the depth-prior weight to $\gamma \in \{0, 5\}$ for every method and report both; the LEMS authors instead tune hyperparameters per cell, which gives stronger KFAC-SVD results at ratio 0.7 than our fixed-$\gamma$ runs. The budget is 90% / 70% of the parameters of all attention and MLP projections; no layer is left dense.

### 3.1 Comparison to State-of-the-art Low-Rank SVD Methods

We evaluate **Complete-SVD** against state-of-the-art SVD-based baselines on Mistral-7B, Llama 3-8B and Qwen3-8B at compression ratios of 0.9 and 0.7 (Table 1). For a fair comparison, KFAC-SVD, IO-SVD, **KL-SVD** and **Complete-SVD** share the same uniform rank allocation and covariance regularization procedure (Appendix F). On the calibration corpus (WikiText-2), **Complete-SVD** remains within 1.3 perplexity points of the stronger of KFAC-SVD and IO-SVD in every configuration. The consistent advantage appears off the calibration distribution: **Complete-SVD** achieves the lowest C4 and PTB perplexity in all six configurations. On Llama 3-8B and Qwen3-8B, it reduces C4 perplexity relative to the stronger of these two baselines by 1–3% at ratio 0.9 and 5–15% at ratio 0.7, and PTB perplexity by 16–44%. Average zero-shot accuracy remains comparable, ranging from 1.4 percentage points below to 0.9 point above the stronger baseline. Under generative evaluation (Table 2), **Complete-SVD** outperforms both KFAC-SVD and IO-SVD on all three models at ratios 0.9 and 0.8. To verify that the perplexity gains are not specific to WikiText-2 calibration, Table 4 repeats the comparison on Llama 3-8B with C4 as the calibration set. **Complete-SVD** slightly improves on the stronger baseline on C4 at ratio 0.9 and reduces its perplexity by 20% at ratio 0.7, while reducing perplexity by 13–26% on WikiText-2 and 27–43% on PTB.

**Comparison with KL-SVD.** Comparing **Complete-SVD** with **KL-SVD** further isolates the effect of the compression objective, as both methods also share the probe-based estimation framework. Relative to **KL-SVD**, **Complete-SVD** has 4–6% higher WikiText-2 perplexity in five of the six configurations and 0.4–1.4 percentage points lower average zero-shot accuracy in all six. Nevertheless, **Complete-SVD** achieves lower C4 and PTB perplexity in every configuration, with reductions of up to 7% and 28%, respectively. These results suggest that the **Complete-SVD** objective better preserves performance beyond the calibration corpus, at the cost of slightly higher calibra-tion perplexity and lower zero-shot accuracy. With C4 calibration (Table 4) on Llama 3-8B, **KL-SVD** attains slightly lower C4 perplexity at ratio 0.9 (15.5 against 15.7), but **Complete-SVD** reduces WikiText-2 perplexity by 16% and PTB perplexity by 5%. At ratio 0.7, **Complete-SVD** is lower on all three corpora, by 15% on WikiText-2, 1% on C4 and 17% on PTB. As with WikiText-2 calibration, **KL-SVD** attains higher average zero-shot accuracy at both ratios.

| CR | Method | Mistral-7B | | | | Llama 3-8B | | | | Qwen3-8B | | | |
|---|---|---|---|---|---|---|---|---|---|---|---|---|---|
| | | Wiki↓ | C4↓ | PTB↓ | Acc↑ | Wiki↓ | C4↓ | PTB↓ | Acc↑ | Wiki↓ | C4↓ | PTB↓ | Acc↑ |
| | Baseline | 5.3 | 8.1 | 27.7 | 64.0 | 6.1 | 9.5 | 9.9 | 63.3 | 9.7 | 15.5 | 15.4 | 62.0 |
| 0.9 | FWSVD | 9.5 | 13.8 | 58.4 | 56.9 | 42.1 | 60.0 | 69.8 | 48.9 | 17.0 | 24.8 | 31.5 | 53.8 |
| | ASVD | 9.1 | 13.4 | 49.1 | **57.3** | 65.1 | 77.6 | 207.7 | 47.9 | 20.4 | 30.0 | 36.4 | 51.2 |
| | SVD-LLM | 6.5 | 15.1 | 66.1 | 56.8 | 10.1 | 39.6 | 49.9 | 52.3 | 12.5 | 27.4 | 36.7 | 56.0 |
| | DOBI-SVD | 7.1 | 15.2 | 63.6 | 55.2 | 11.2 | 34.8 | 42.4 | 53.0 | 13.5 | 28.9 | 35.8 | 55.5 |
| | GFWSVD | 31.7 | 65.1 | 305.7 | 40.7 | 2570 | 3828 | 8492 | 33.4 | 171.5 | 350.5 | 765.6 | 46.7 |
| | KFAC-SVD | 6.2 | 13.1 | 44.0 | 56.9 | 8.8 | 25.7 | 32.6 | 54.5 | 11.5 | 23.8 | 28.0 | 57.1 |
| | IO-SVD | **6.1** | 12.8 | 40.0 | 56.9 | 8.4 | 23.2 | 30.9 | 56.2 | **11.0** | 23.7 | 28.9 | **57.9** |
| | **KL-SVD** | **6.1** | 12.8 | 40.1 | **57.3** | **8.3** | 22.8 | 27.1 | **56.9** | 11.2 | 24.6 | 29.1 | 57.7 |
| | **Complete-SVD** | 6.4 | **12.7** | **39.9** | 56.9 | 8.7 | **22.4** | **23.0** | 55.8 | 11.7 | **23.4** | **23.4** | 56.7 |
| 0.7 | FWSVD | 34.8 | 50.3 | 215.1 | 44.2 | 716.4 | 837.5 | 1940 | 33.4 | 41.4 | 68.8 | 90.7 | 42.7 |
| | ASVD | 28.2 | 42.9 | 163.6 | 46.1 | 10989 | 9400 | 36033 | 33.0 | 70.7 | 97.0 | 131.5 | 43.1 |
| | SVD-LLM | 11.0 | 42.6 | 169.5 | 44.4 | 34.6 | 376.0 | 600.9 | 37.7 | 17.1 | 63.1 | 136.2 | 45.8 |
| | DOBI-SVD | 12.4 | 39.7 | 170.8 | 42.9 | 31.5 | 500.1 | 223.7 | 38.5 | 18.5 | 63.8 | 94.3 | 46.2 |
| | GFWSVD | 2550 | 3404 | 5356 | 32.3 | 41799 | 53671 | 72222 | 31.8 | 21685 | 27627 | 50815 | 34.1 |
| | KFAC-SVD | 9.3 | 34.6 | 110.7 | 45.8 | 19.4 | 151.3 | 253.5 | 40.7 | 14.7 | 50.0 | 94.0 | 46.8 |
| | IO-SVD | **8.8** | 33.1 | 97.8 | 45.6 | 16.8 | 111.8 | 219.2 | 43.3 | 14.4 | 58.7 | 120.2 | **49.2** |
| | **KL-SVD** | 10.3 | 34.7 | 98.8 | **46.6** | **16.0** | 103.0 | 172.7 | **45.6** | **14.2** | 49.9 | 82.8 | 48.6 |
| | **Complete-SVD** | 10.1 | **32.9** | **90.7** | 45.4 | 16.8 | **95.3** | **123.7** | 44.2 | 15.0 | **47.5** | **67.7** | 47.8 |

Table 1: Compression results on Mistral-7B, Llama 3-8B and Qwen3-8B with uniform rank allocation. Results for IO-SVD are reproduced by us; all other baselines are reported from Thoma et al. (2026). Accuracy (Acc) reported as average zero-shot accuracy value on 6 downstream datasets.

| CR | Method | Mistral 7B | Qwen3 8B | Llama 2 7B |
|---|---|---|---|---|
| | Baseline | 16.8 | 24.4 | 30.3 |
| 0.9 | KFAC-SVD | 12.4 | 17.2 | 20.2 |
| | IO-SVD | 13.9 | 12.2 | 22.4 |
| | **KL-SVD** | 12.4 | 17.7 | **27.7** |
| | **Complete-SVD** | **17.4** | **21.3** | 26.6 |
| 0.8 | KFAC-SVD | 11.8 | 14.6 | 22.2 |
| | IO-SVD | 7.4 | 10.5 | 15.3 |
| | **KL-SVD** | 11.3 | 13.3 | 23.5 |
| | **Complete-SVD** | **12.9** | **18.1** | **25.3** |

Table 2: IFEval (0-shot, prompt-level strict accuracy, without applying a chat template) under low-rank compression at compression ratios (CR) 0.9 and 0.8. Llama 2-7B is the Chat variant.

**Sensitivity-aware rank allocation.** Table 3 reports the output-aware methods under LEMS. The allocator decides how many directions each layer keeps, the objective which ones. At ratio 0.9, allocation alone brings every method close to the uncompressed model (Llama 3-8B C4: 11.4 against 9.5, versus 22.4 with uniform ranks), leaving little for the objective to change: **Complete-SVD** and **KL-SVD** differ by at most 0.5 perplexity points. At ratio 0.7, the objective matters again: **Complete-SVD** achieves the lowest C4 and PTB perplexity on all three models and for both values of $\gamma$, and its PTB reduction over the strongest competitor (9–34%) is at least as large as with uniform ranks (7–28%). The ordering holds for both $\gamma$ although $\gamma$ changes perplexity by up to a factor of two, so it does not hinge on the allocator's tuning. As under uniform ranks, **KL-SVD** keeps the higher zero-shot accuracy.

**Comparison with pruning methods.** Following the setup of Dobi-SVD (Wang et al., 2025b), we also compare **Complete-SVD** with post-training pruning methods (Table 5). At 10.1 GB, LLM-Pruner and Wanda-SP reach WikiText-2 perplexities of 10.5 and 12.1, against 7.5 for

| CR | Method | $\gamma$ | Mistral-7B | | | | Llama 3-8B | | | | Qwen3-8B | | | |
|---|---|---|---|---|---|---|---|---|---|---|---|---|---|---|
| | | | Wiki↓ | C4↓ | PTB↓ | Acc↑ | Wiki↓ | C4↓ | PTB↓ | Acc↑ | Wiki↓ | C4↓ | PTB↓ | Acc↑ |
| | Baseline | - | 5.3 | 8.1 | 27.7 | 64.0 | 6.1 | 9.5 | 9.9 | 63.3 | 9.7 | 15.5 | 15.4 | 62.0 |
| 0.9 | KFAC-SVD | 0 | **5.4** | 8.6 | 26.8 | 62.3 | 6.7 | 11.6 | 11.5 | 60.5 | **9.9** | **16.2** | **15.9** | **61.9** |
| | **KL-SVD** | | **5.4** | **8.5** | **26.5** | **62.4** | **6.6** | **11.4** | 11.2 | **61.0** | **9.9** | **16.2** | 16.0 | 61.8 |
| | **Complete-SVD** | | **5.4** | 8.6 | 27.0 | 62.2 | **6.6** | **11.4** | **11.1** | **61.0** | **9.9** | **16.2** | 16.0 | 61.8 |
| | KFAC-SVD | 5 | 5.5 | 8.8 | 29.3 | 62.3 | 7.1 | 12.6 | 12.9 | 62.0 | 10.1 | 16.7 | 16.6 | 61.1 |
| | **KL-SVD** | | **5.4** | **8.7** | **27.2** | **62.6** | **7.0** | 12.3 | 12.0 | **62.1** | 10.1 | 16.7 | 16.8 | **61.3** |
| | **Complete-SVD** | | **5.4** | **8.7** | 27.4 | 62.3 | **7.0** | **12.2** | **11.6** | 61.3 | **10.0** | **16.6** | **16.5** | 60.9 |
| 0.7 | KFAC-SVD | 0 | 8.5 | 30.3 | 100.9 | 47.6 | 15.6 | 82.8 | 162.5 | 45.8 | 12.2 | 33.7 | 48.0 | 52.1 |
| | **KL-SVD** | | **8.2** | 28.8 | 83.7 | **48.3** | **13.1** | 65.0 | 117.5 | **47.3** | 12.0 | 33.2 | 50.3 | **53.5** |
| | **Complete-SVD** | | 8.7 | **27.9** | **74.2** | 46.8 | 13.4 | **60.9** | **77.3** | 46.0 | **11.9** | **31.1** | **36.3** | 52.8 |
| | KFAC-SVD | 5 | 7.6 | 20.1 | 65.2 | 52.0 | 11.8 | 38.1 | 88.2 | 50.0 | **13.0** | 34.8 | 66.6 | **52.8** |
| | **KL-SVD** | | **7.3** | 18.4 | 56.5 | **53.1** | **11.0** | 34.2 | 73.1 | **51.5** | 13.4 | 35.7 | 69.0 | **52.8** |
| | **Complete-SVD** | | 7.5 | **18.0** | **51.6** | 51.6 | 11.1 | **32.5** | **50.8** | 50.6 | 13.2 | **33.2** | **47.7** | 51.8 |

Table 3: Compression results on Mistral-7B, Llama 3-8B and Qwen3-8B with LEMS rank allocation at matched budget and fixed hyperparameters.

**Complete-SVD**; at 6.8 GB, both exceed 249, whereas **Complete-SVD** attains 19.8. Under the remapping storage of Dobi-SVD (marked *, Appendix E), **Complete-SVD*** outperforms Dobi-SVD* at every memory budget, although it uses uniform ranks whereas Dobi-SVD* uses its loss-aware rank allocation.

| *Memory (GB)* | 10.1 | 7.7 | 6.8 |
|---|---|---|---|
| | | Wiki↓ | |
| LLM-Pruner | 10.5 | 46.3 | 253.1 |
| Wanda-SP | 12.1 | 38.6 | 249.2 |
| **Complete-SVD** | **7.5** | **13.3** | **19.8** |
| Dobi-SVD* | 5.9 | 7.9 | 9.5 |
| **Complete-SVD*** | **5.8** | **7.0** | **8.4** |

Table 5: Perplexity results (Wiki-2) of **Complete-SVD** versus state-of-the-art pruning methods and Dobi-SVD on Llama 2-7B (Base). Methods marked with * use remapping storage.

### 3.2 Ablation and Sensitivity Studies

**Ablation Study on Covariance Components.** We analyze the individual contributions of $\Sigma_{\text{grad}}$ and $\Sigma_{\text{input}}$ to the SVD compression objective on Llama 2-7B (Base). As shown in Table 6, while each matrix individually improves over standard SVD, their combination consistently yields the best performance across all compression ratios. At 50% compression ratio, **Complete-SVD** achieves substantially lower perplexity than using $\Sigma_{\text{input}}$ alone or $\Sigma_{\text{grad}}$ alone, demonstrating that each component contributes complementary information to the functional norm.

| CR | Method | Llama 3-8B | | | |
|---|---|---|---|---|---|
| | | Wiki ↓ | C4 ↓ | PTB ↓ | Acc ↑ |
| | Baseline | 6.1 | 9.5 | 9.9 | 63.3 |
| 0.9 | KFAC-SVD | 17.7 | 15.9 | 23.6 | 55.9 |
| | IO-SVD | 14.7 | 16.1 | 25.8 | 56.8 |
| | **KL-SVD** | 15.2 | **15.5** | 18.3 | **57.6** |
| | **Complete-SVD** | **12.8** | 15.7 | **17.3** | 56.8 |
| 0.7 | KFAC-SVD | 89.0 | 47.3 | 158.6 | 43.6 |
| | IO-SVD | 69.5 | 45.5 | 169.0 | 44.9 |
| | **KL-SVD** | 60.1 | 36.7 | 108.5 | **46.4** |
| | **Complete-SVD** | **51.2** | **36.2** | **89.9** | 45.7 |

Table 4: Compression results of output-aware methods on Llama 3-8B (WikiText-2, C4, and PTB perplexity, zero-shot accuracy) when C4 dataset is used for calibration with 256 samples and 2048 token sequence.

**Robustness to Calibration Data.** We assess the sensitivity of **Complete-SVD** to its random draws and calibration budget (Figure 2). At the reference budget ($N = 256$ sequences, $K = 64$ probes), varying the calibration seed (resampling the sequences) give WikiText-2 perplexities in $\{7.53, 7.54, 7.57\}$ (std $s_{\text{cal}} = 0.02$), and the estimator seed (redrawing the probes $\Psi$ and signs $\varepsilon$ of $\tilde{\Sigma}_{\text{grad}}$) give $\{7.49, 7.53, 7.55\}$ (std $s_{\text{est}} = 0.03$). Reducing $N$ from 256 to 32 slightly raises perplexity monotonically from 7.53 to 7.61, and varying $K$ over $\{64, 128, 256\}$ changes it by at most 0.03 (within noise). We use $N = 256$ and $K = 64$ as a compromise between cost and precision. Table 7 compares covariance matrices estimated on WikiText-2, our default, against C4:

| CR | Method | $C_{\text{input}}$ | $C_{\text{grad}}$ | Wiki ↓ |
|---|---|---|---|---|
| | Baseline | N/A | N/A | 5.47 |
| 0.8 | SVD | Id | Id | 19 438.3 |
| | Gradient only | Id | $\Sigma_{\text{grad}}$ | 21.8 |
| | Forward only | $\Sigma_{\text{input}}$ | Id | 8.5 |
| | **Complete-SVD** | $\Sigma_{\text{input}}$ | $\Sigma_{\text{grad}}$ | **7.5** |
| 0.7 | SVD | Id | Id | 31 062.3 |
| | Gradient only | Id | $\Sigma_{\text{grad}}$ | 81.9 |
| | Forward only | $\Sigma_{\text{input}}$ | Id | 10.9 |
| | **Complete-SVD** | $\Sigma_{\text{input}}$ | $\Sigma_{\text{grad}}$ | **9.3** |
| 0.6 | SVD | Id | Id | 44 494.7 |
| | Gradient only | Id | $\Sigma_{\text{grad}}$ | 290.0 |
| | Forward only | $\Sigma_{\text{input}}$ | Id | 17.3 |
| | **Complete-SVD** | $\Sigma_{\text{input}}$ | $\Sigma_{\text{grad}}$ | **13.3** |
| 0.5 | SVD | Id | Id | 53 670.8 |
| | Gradient only | Id | $\Sigma_{\text{grad}}$ | 2211.0 |
| | Forward only | $\Sigma_{\text{input}}$ | Id | 37.9 |
| | **Complete-SVD** | $\Sigma_{\text{input}}$ | $\Sigma_{\text{grad}}$ | **22.4** |

Table 6: WikiText-2 perplexity of Llama 2-7B (Base) when optimizing $\left\| C_{\text{input}}^{\frac{1}{2}} \mathrm{d}W C_{\text{grad}}^{\frac{1}{2}} \right\|_F$.

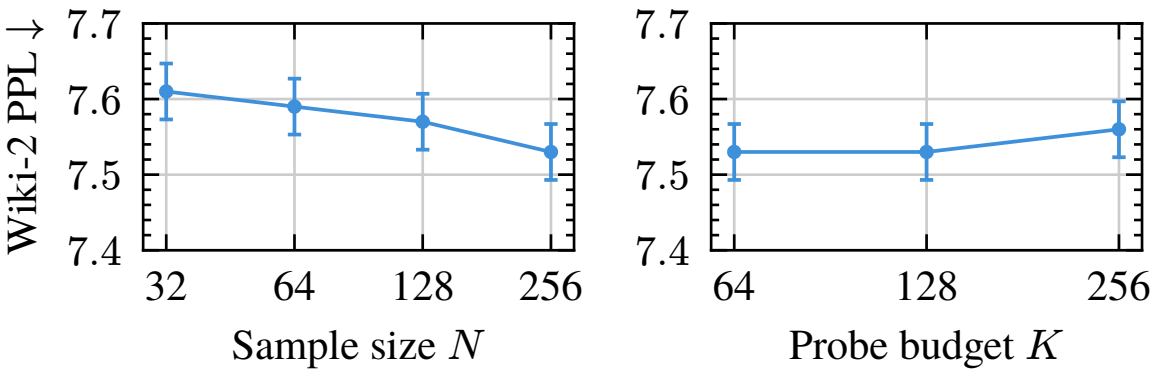

Figure 2: WikiText-2 perplexity of **Complete-SVD** on Llama 2-7B (Base) at 80% compression ratio versus the sample size $N$ (left, $K = 64$) and the probe budget $K$ (right, $N = 256$). Error bars show $\pm\sigma$, with $\sigma = \sqrt{s_{\text{cal}}^2 + s_{\text{est}}^2} \approx 0.04$ the single-run standard deviation estimated from seed repeats at the reference budget.

| **Cal. Set** | Wiki ↓ | C4 ↓ | PTB ↓ | Acc. ↑ |
|---|---|---|---|---|
| Wiki-2 | **7.5** | 15.2 | 56.9 | 50.0 |
| C4 | 9.9 | **11.6** | **56.0** | **51.3** |

Table 7: Effect of the calibration set (WikiText-2 vs. C4) on **Complete-SVD** applied to Llama 2-7B (Base) at 80% compression ratio. Average accuracy is computed over 6 downstream datasets.

each set favors its own distribution in perplexity, with comparable downstream accuracy.

**Compatibility with quantization.** We verify this by applying 3-bit weight-only GPTQ quantization (W3A16) to Llama 2-7B compressed by **Complete-SVD** at several ratios (Table 8). At ratio 0.8, quantization reduces the memory footprint from 10.1 GB to 2.5 GB at a WikiText-2 perplexity of 9.2; at ratio 0.6, the model occupies 1.9 GB, 14% of the original FP16 size. The low-rank factors produced by **Complete-SVD** therefore remain compatible with quantization, and the two techniques can be combined for deployment on memory-constrained devices.

| **CR** | **Method** | Wiki↓ | **Mem**↓ |
|---|---|---|---|
| 0.8 | **Complete-SVD** | 7.5 | 10.1 GB |
| | **Complete-SVD** + GPTQ W3A16 | 9.2 | 2.5 GB |
| 0.6 | **Complete-SVD** | 13.3 | 7.7 GB |
| | **Complete-SVD** + GPTQ W3A16 | 16.7 | 1.9 GB |

Table 8: Perplexity (Wiki) and model size for **Complete-SVD** combined with GPTQ quantization on Llama 2-7B (Base). Memory reports the model parameter storage after compression.

## 4 Conclusion

We introduced **Complete-SVD**, a post-training low-rank compression method that minimizes the $L_2$ distance between the logits of the original and compressed networks over the data distribution. We make the Gram matrix of logit Jacobians, $\Sigma_{\text{grad}}$, tractable by exactly reducing its vocabulary dimension to the hidden dimension and estimating it without bias using $K$ backward passes per sequence, independent of vocabulary size and sequence length. As a controlled counterpart, we proposed **KL-SVD** which replaces $L_2$ distance with KL divergence between predictive distributions, and uses an unbiased estimator of the true Fisher matrix.

Across Mistral-7B, Llama 3-8B and Qwen3-8B, **Complete-SVD** achieves the lowest C4 and PTB perplexity among all compared methods, including **KL-SVD**, under WikiText-2 calibration, with covariance regularization held fixed across the output-aware methods. This cross-corpus advantage holds under both uniform and sensitivity-aware (LEMS) rank allocation, the latter at ratio 0.7, and persists with C4 calibration. **Complete-SVD** also improves IFEval instruction-following accuracy over KFAC-SVD and IO-SVD in every tested setting, and over **KL-SVD** in all but one setting. **KL-SVD**, in turn, attains equal or higher average zero-shot accuracy in every configuration. **Complete-SVD** requires no fine-tuning and is compatible with post-training quantization. These results suggest that, for post-training low-rank compression, matching the logits of the original model is a simple and effective alternative to preserving either the loss on the observed next token or the predicted distribution.

## 5 Limitations

Firstly, our evaluation is limited to models in the 7–8B parameter range. Evaluating larger models would provide stronger evidence that the observed behavior of the different output-aware methods persists at scale. Secondly, as in most post-training SVD-based methods, each layer is compressed using statistics collected from the original model, so the errors introduced by previously compressed layers are not taken into account. AA-SVD (Sinha and Fleuret, 2026) mitigates this by combining an activation-aware factorization with a gradient-based block-wise refinement on the activations of the partially compressed network, at the cost of no longer being training-free. Reducing calibration cost and estimator variance without sacrificing the full functional objective remains an important direction for future work. Finally, the complete functional norm is not specific to low-rank factorization. GPTQ (Frantar et al., 2023) and SparseGPT (Frantar and Alistarh, 2023) minimize $\left\|\Sigma_{\text{input}}^{\frac{1}{2}} \mathrm{d}W\right\|_F$, that is, the objective of Theorem 1 with $\Sigma_{\text{grad}} = \mathrm{Id}$. Substituting $\Sigma_{\text{grad}}$ yields an output-aware objective for quantization and pruning, although the closed-form SVD solution no longer applies and a dedicated solver is required, as in YAQA (Tseng et al., 2026) for the KL objective.

## A Proof on Complete Functional Norm

**Theorem 1.** *With the notation introduced in Section 2.1, let* $\Sigma_{\text{input}} := \mathbb{E}_{x\sim\mathcal{D}}[a(x)(a(x))^\top]$ *denote the input covariance and* $\Sigma_{\text{grad}} := \mathbb{E}_{x\sim\mathcal{D}}\left[\left(\frac{\partial f_W}{\partial s}(x)\right)^\top \frac{\partial f_W}{\partial s}(x)\right]$ *the Gram matrix of the output Jacobians at the compressed layer. Then:*

(i) Exact first-order expansion. *Let* $G_W := \mathbb{E}_{x\sim\mathcal{D}}\left[\left(\frac{\partial f_W(x)}{\partial W}\right)^\top \frac{\partial f_W(x)}{\partial W}\right] \in \mathbb{R}^{n_s n_a \times n_s n_a}$. *Then*

$$\|f_W - f_{\widehat{W}}\| = \left\| G_W^{\frac{1}{2}} \overrightarrow{\mathrm{d}W} \right\|_F + o(\|\mathrm{d}W\|_F). \tag{5}$$

(ii) Kronecker-factored approximation. *Under the K-FAC structural approximation* $G_W \approx \Sigma_{\text{input}} \otimes \Sigma_{\text{grad}}$, *the right-hand side of Equation 5 admits the closed-form approximation*

$$\left\| G_W^{\frac{1}{2}} \overrightarrow{\mathrm{d}W} \right\|_F \approx \left\| \Sigma_{\text{input}}^{\frac{1}{2}} \mathrm{d}W \Sigma_{\text{grad}}^{\frac{1}{2}} \right\|_F. \tag{6}$$

*Proof. Step 1 (exact first-order expansion).* Writing the first-order Taylor development of $f_W$ around $W$,

$$f_{\widehat{W}} - f_W = \frac{\partial f_W}{\partial W}\overrightarrow{\mathrm{d}W} + o(\|\mathrm{d}W\|_F). \tag{7}$$

Taking norms and expanding the inner product (full derivation in Appendix B.2),

$$\|f_W - f_{\widehat{W}}\|^2 = \left\| \frac{\partial f_W}{\partial W}\overrightarrow{\mathrm{d}W} \right\|_F^2 + o(\|\mathrm{d}W\|_F^2) \tag{8}$$

$$= \left(\overrightarrow{\mathrm{d}W}\right)^\top G_W \overrightarrow{\mathrm{d}W} + o(\|\mathrm{d}W\|_F^2) \tag{9}$$

where $G_W := \left(\frac{\partial f_W}{\partial W}\right)^\top \frac{\partial f_W}{\partial W}$. Taking square roots yields Equation 5.

*Step 2 (K-FAC approximation).* The matrix $G_W$ has size $n_s n_a \times n_s n_a$ and is intractable to store or factorize at LLM scale. We invoke the K-FAC structural approximation (Dangel et al., 2025),

$$G_W \approx \Sigma_{\text{input}} \otimes \Sigma_{\text{grad}}, \tag{10}$$

where $\Sigma_{\text{input}} := \mathbb{E}_{x\sim\mathcal{D}}(a(x)(a(x))^\top)$ and $\Sigma_{\text{grad}} := \mathbb{E}_{x\sim\mathcal{D}}\left(\left(\frac{\partial f_W}{\partial s}(x)\right)^\top \frac{\partial f_W}{\partial s}(x)\right)$. This is a structural approximation: see Remark 2.

Combining the Kronecker identities $(X \otimes Y)^{\frac{1}{2}} = X^{\frac{1}{2}} \otimes Y^{\frac{1}{2}}$ and $(X \otimes Y)\,\mathrm{vec}(Z) = \mathrm{vec}(XZY^\top)$ with Equation 10 gives

$$\left\| G_W^{\frac{1}{2}} \overrightarrow{\mathrm{d}W} \right\|_F \approx \left\| (\Sigma_{\text{input}} \otimes \Sigma_{\text{grad}})^{\frac{1}{2}} \overrightarrow{\mathrm{d}W} \right\|_F \tag{11}$$

$$= \left\| \Sigma_{\text{input}}^{\frac{1}{2}} \mathrm{d}W \Sigma_{\text{grad}}^{\frac{1}{2}} \right\|_F,$$

which is Equation 6. □

**Remark 2.** *The two approximation symbols in Theorem 1 are not interchangeable. The remainder in Equation 5 is* $o(\|\mathrm{d}W\|_F)$ *and vanishes as* $\mathrm{d}W \to 0$. *The K-FAC step in Equation 6, by contrast, replaces the full Gram matrix* $G_W$ *by its Kronecker-factored approximation; the resulting structural error* $\varepsilon_{\text{KFAC}} := \|G_W - \Sigma_{\text{input}} \otimes \Sigma_{\text{grad}}\|_F$ *is independent of* $\mathrm{d}W$ *and does not vanish in any limit. We adopt the K-FAC approximation because it admits a closed-form SVD solution and has been empirically validated as a tight approximation of the layer-wise curvature for fully-connected layers (Dangel et al., 2025; Eschenhagen et al., 2023); we do not claim it is asymptotically tight in* $\mathrm{d}W$.

## B Reconstruction Error

The first idea to find a good $\widehat{W}$ from $W$ is to minimize $\left\|\widehat{W} - W\right\|_F$ which can be done by computing the SVD of $W$ and keeping the top $k$ singular values. However, this does not take into account the effect of the change in weight on the loss. Hence, we can aim instead to minimize the change in loss. Writing $\mathrm{d}W = \widehat{W} - W$ and $\overrightarrow{\mathrm{d}W}$ the vectorization of $\mathrm{d}W$ we have the following first-order Taylor expansion of the loss around $W$:

$$\mathcal{L}(f_{\widehat{W}}) - \mathcal{L}(f_W) \tag{12}$$

$$= \left\langle \nabla_W \mathcal{L}(f_W), \overrightarrow{\mathrm{d}W} \right\rangle_2 + o(\|\mathrm{d}W\|_F)$$

This suggests choosing $\widehat{W}$ in order to minimize $\left\langle \nabla_W \mathcal{L}(f_W), \overrightarrow{\mathrm{d}W} \right\rangle$. However this would lead to only approximate $f_W$ on the correct label and does not give a rich feedback. As shown for Knowledge distillation (Hinton et al., 2015), it is better to copy the full distribution and to have a richer feedback. In addition, as we suppose that $f_W$ is a local minimizer (as it is an already trained function), we can expect that $\nabla_W \mathcal{L}(f_W) \approx 0$ and that the gradient information is limited.

Hence we can either directly minimize the functional distance between the output of $f_{\widehat{W}}$ and $f_W$:

$$\|f_W - f_{\widehat{W}}\|^2 \tag{13}$$

$$= \mathbb{E}_{x\sim\mathcal{D}}\left(\left\|f_W(x) - f_{\widehat{W}}(x)\right\|_F^2\right)$$

or minimize the expected KL divergence between the output of $f_{\widehat{W}}$ and $f_W$:

$$\mathrm{KL}\left(f_W, f_{\widehat{W}}\right) \tag{14}$$
$$:= \mathbb{E}_{x\sim\mathcal{D}}\left(\mathsf{KL}\left(\mathrm{Softmax}(f_W(x)),\right.\right.$$
$$\left.\left.\mathrm{Softmax}\left(f_{\widehat{W}}(x)\right)\right)\right)$$

two objectives that capture the full output distribution and not only the value at the correct label.

We show now how we can optimize those objectives efficiently.

### B.1 Norm Optimization

What we can show is that minimizing Equation 12, Equation 14 or Equation 13 can be done with similar techniques. Each of those objectives can be approximated by a quadratic form of $\overrightarrow{\mathrm{d}W}$ with a curvature matrix that can be approximated using the Kronecker-factored approximation (Dangel et al., 2025) as a Kronecker product of two matrices $C_{\text{input}}$ and $C_{\text{grad}}$.

### B.2 Hilbertian Norm Minimization

Writing the first order development of $f_W$ around $W$ we have:

$$f_{\widehat{W}} - f_W = \frac{\partial f_W}{\partial W}\overrightarrow{\mathrm{d}W} + o(\|\mathrm{d}W\|_F) \tag{15}$$

Hence:

$$\left\|f_W - f_{\widehat{W}}\right\|^2 = \left\|\frac{\partial f_W}{\partial W}\overrightarrow{\mathrm{d}W}\right\|^2 \tag{16}$$
$$+\left\langle\frac{\partial f_W}{\partial W}\overrightarrow{\mathrm{d}W}, o(\|\mathrm{d}W\|_F)\right\rangle + o(\|\mathrm{d}W\|_F)^2$$
$$= \left\|\frac{\partial f_W}{\partial W}\overrightarrow{\mathrm{d}W}\right\|^2 \tag{17}$$
$$+\left\langle\frac{\partial f_W}{\partial W}, o(\|\mathrm{d}W\|_F)\overrightarrow{\mathrm{d}W}^\top\right\rangle + o(\|\mathrm{d}W\|_F^2)$$
$$= \left\|\left(\left(\frac{\partial f_W}{\partial W}\right)^\top\frac{\partial f_W}{\partial W}\right)^{\frac{1}{2}}\overrightarrow{\mathrm{d}W}\right\|_F^2 \tag{18}$$
$$+o(\|\mathrm{d}W\|_F^2)$$

Then taking the square-root we have:

$$\|f_W - f_{\widehat{W}}\| = \left\|\left(\left(\frac{\partial f_W}{\partial W}\right)^\top\frac{\partial f_W}{\partial W}\right)^{\frac{1}{2}}\overrightarrow{\mathrm{d}W}\right\|_F \tag{19}$$
$$+o(\|\mathrm{d}W\|_F)$$

This is still expensive to compute, as for fully-connected layers, $G_W := \left(\frac{\partial f_W}{\partial W}\right)^\top\frac{\partial f_W}{\partial W}$ is a matrix of size $n_s n_a \times n_s n_a$. To further reduce the computational cost, we can use the Kronecker-factored approximation (Dangel et al., 2025) to approximate $G_W$ by the Kronecker product of two matrices $\Sigma_{\text{input}} := \mathbb{E}_{x\sim\mathcal{D}}\left(a(x)\left(a(x)\right)^\top\right)$ and $\Sigma_{\text{grad}} := \mathbb{E}_{x\sim\mathcal{D}}\left(\left(\frac{\partial f_W}{\partial s}(x)\right)^\top\frac{\partial f_W}{\partial s}(x)\right)$ : $G_W \approx \Sigma_{\text{input}} \otimes \Sigma_{\text{grad}}$.

### B.3 Direct Loss Minimization

We can minimize the absolute value as we can suppose that we won't get a lower loss after compression. Rewriting the objective:

$$\left|\left\langle\nabla_W\mathcal{L}(f_W), \overrightarrow{\mathrm{d}W}\right\rangle_2\right| \tag{20}$$
$$= \left\|\nabla_W\mathcal{L}(f_W)^\top\overrightarrow{\mathrm{d}W}\right\|_F$$
$$= \left\|\left(\nabla_W\mathcal{L}(f_W)(\nabla_W\mathcal{L}(f_W))^\top\right)^{\frac{1}{2}}\overrightarrow{\mathrm{d}W}\right\|_F$$

### B.4 KL Minimization

A classical result in information theory is that the change in KL divergence can be approximated by a quadratic form of the Fisher information matrix of the distribution $f_W(x)$ with respect to the parameters $W$ noted $F_W(x)$:

$$\mathrm{KL}\left(\mathrm{Softmax}(f_W(x)), \mathrm{Softmax}\left(f_{\widehat{W}}(x)\right)\right)$$
$$= \frac{1}{2}\left(\overrightarrow{\mathrm{d}W}\right)^\top F_W(x)\overrightarrow{\mathrm{d}W} + o(\|\mathrm{d}W\|_F^2) \tag{21}$$

where the Fisher information matrix is defined for $p_W(x) := \mathrm{Softmax}(f_W(x))$ the label distribution predicted by the model parameterized by $W$ as:

$$F_W(x) = \mathbb{E}_{y\sim p_W(x)}\Big[\nabla_W \log p_W(y|x) \times$$
$$\times\, (\nabla_W \log p_W(y|x))^\top\Big]. \tag{22}$$

Hence taking the expectation over the data distribution, we have with $F_W := \mathbb{E}_{x\sim\mathcal{D}}(F_W(x))$

$$\mathrm{KL}\left(f_W, f_{\widehat{W}}\right) \tag{23}$$

$$= \frac{1}{2}\left(\overrightarrow{\mathrm{d}W}\right)^\top F_W \overrightarrow{\mathrm{d}W} + o(\|\mathrm{d}W\|_F^2)$$

$$\approx \frac{1}{2}\left\| F_W^{\frac{1}{2}} \overrightarrow{\mathrm{d}W} \right\|_F^2$$

and as shown by Dangel et al. (2025), $F_W$ can be approximated by the Kronecker product of two matrices $\Sigma_{\text{input}}$ and $F_s$: $F_W \approx \Sigma_{\text{input}} \otimes F_s$ where $F_s$ is the expected Fisher information matrix of the distribution $f_W(x)$ with respect to values $s(x)$.

If we note $p_W(x) := \mathrm{Softmax}(f_W(x))$ we have:

$$F_s = \mathbb{E}_{x\sim\mathcal{D}}[F_s(x)] \tag{24}$$

$$= \mathbb{E}_{x\sim\mathcal{D}}\Big[\mathbb{E}_{y\sim p_W(x)}\Big[\nabla_{s(x)} \log p_W(y|x) \times \times \left(\nabla_{s(x)} \log p_W(y|x)\right)^\top\Big]\Big] \tag{25}$$

### B.4.1 Optimization of the Quadratic Forms

All of above objectives can be formulated as minimizing $\left\|C^{\frac{1}{2}}\overrightarrow{\mathrm{d}W}\right\|_F$ where $C$ is a conditioning matrix that can be approximated by the Kronecker product of two matrices $C_{\text{input}}$ and $C_{\text{grad}}$. Using the properties of the Kronecker product, $(X \otimes Y)^{\frac{1}{2}} = X^{\frac{1}{2}} \otimes Y^{\frac{1}{2}}$ and $(X \otimes Y)\mathrm{vec}(Z) = \mathrm{vec}(XZY^\top)$ (where vec is the row-wise vectorization), we have:

$$d(f_W, f_{\widehat{W}}) \approx \left\| (C_{\text{input}} \otimes C_{\text{grad}})^{\frac{1}{2}} \overrightarrow{\mathrm{d}W} \right\|_F$$

$$= \left\| C_{\text{input}}^{\frac{1}{2}} \mathrm{d}W C_{\text{grad}}^{\frac{1}{2}} \right\|_F \tag{26}$$

This final optimization problem can be solved efficiently in closed-form using the SVD.

| Method | $C_{\text{input}}$ | $C_{\text{grad}}$ |
|---|---|---|
| Empirical Fisher curvature (Chekalina et al., 2026) | $\Sigma_{\text{input}}$ | $E_s$ |
| KL curvature | $\Sigma_{\text{input}}$ | $F_s$ |
| L2 curvature | $\Sigma_{\text{input}}$ | $\Sigma_{\text{grad}}$ |

Table 9: Kronecker-factored curvatures of the output-aware objectives. Each objective weights the approximation error of $W$ by $C_{\text{input}} \otimes C_{\text{grad}}$. The input factor $C_{\text{input}} = \Sigma_{\text{input}}$ is common to all methods, which differ only in the output factor $C_{\text{grad}}$.

# C Links between the different curvature matrices

## C.1 Fisher and Empirical Fisher

The empirical Fisher matrix is called "empirical" because it evaluates the gradient at the observed label instead of taking the expectation over the predictive distribution of the model, as the true Fisher matrix does. The full Fisher matrix is defined as:

$$F_W(x) = \mathbb{E}_{y\sim p_W(x)}\Big[\nabla_W \log p_W(y \mid x) \times \times (\nabla_W \log p_W(y \mid x))^\top\Big] \tag{27}$$

a common approximation is to replace the expectation by the value at the correct label $y(x)$. This is a logical choice as the correct label is the most important one for the loss.

If we approximate $p_W(x)$ by a Dirac distribution on the correct label $\delta(y(x))$, we have:

$$F_W(x) \approx \mathbb{E}_{y\sim\delta(y(x))}\Big[\nabla_W \log p_W(y \mid x) \times \times (\nabla_W \log p_W(y \mid x))^\top\Big] \tag{28}$$

$$= \nabla_W \log p_W(y(x) \mid x) \times \times (\nabla_W \log p_W(y(x) \mid x))^\top \tag{29}$$

As the cross-entropy loss is $\ell(x, y) = -\log(p_W(y(x) \mid x)) + \text{constant}$, where $y(x)$ denote the label of $x$, we have:

$$= \nabla_W \ell(f_W(x, y(x)))(\nabla_W \ell(f_W(x, y(x))))^\top) =: E_W(x) \tag{30}$$

$E$ is called the **empirical** Fisher. Similar to the True Fisher, it can be approximated by a Kronecker product of two matrices:

$$E_W \approx \Sigma_{\text{input}} \otimes E_s \tag{31}$$

## C.2 Links in the $C_{\text{grad}}$ Computations

All the $C_{\text{grad}}$ can be defined with a similar structure:

$$\Sigma_{\text{grad}} = \mathbb{E}_{x\sim\mathcal{D}}\left(\left(\frac{\partial f_W}{\partial s}(x)\right)^\top \frac{\partial f_W}{\partial s}(x)\right) \tag{32}$$

For the Fisher using the chain rule (see Dangel et al. (2025) for more details) we have:

$$F_s = \mathbb{E}_{x\sim\mathcal{D}}\Big[\mathbb{E}_{y\sim p_W(x)}\Big[\nabla_{s(x)} \log p_W(y|x) \times$$

$$\times \left(\nabla_{s(x)} \log p_W(y|x)\right)^\top\Big]\Big] \quad (33)$$

$$= \mathbb{E}_{x\sim\mathcal{D}}\left(\left(\frac{\partial f_W}{\partial s}(x)\right)^\top \Big[\mathrm{diag}(p_W(x)) - p_W(x)p_W(x)^\top\Big]\frac{\partial f_W}{\partial s}(x)\right) \quad (34)$$

The same trick can be applied to the empirical Fisher:

$$E_s = \mathbb{E}_{x\sim\mathcal{D}}\Big[\nabla_s\mathcal{L}(f)(x)\nabla_s\mathcal{L}(f)(x)^\top\Big] \quad (35)$$

$$= \mathbb{E}_{x\sim\mathcal{D}}\Big[\left(\tfrac{\partial f_W}{\partial s}(x)\right)^\top\left(p_W(x) - e_y\right) \times \left(p_W(x) - e_y\right)^\top\left(\tfrac{\partial f_W}{\partial s}(x)\right)\Big] \quad (36)$$

So all methods can be formulated as:

$$C_{\text{grad}} = \mathbb{E}_{x\sim\mathcal{D}}\left(\left(\frac{\partial f_W}{\partial s}(x)\right)^\top \mathcal{K}(x,y)\frac{\partial f_W}{\partial s}(x)\right) \quad (37)$$

with:

| Method | $\mathcal{K}(x,y)$ |
|---|---|
| $\Sigma_{\text{grad}}$ ($L_2$ norm) | $I$ |
| $E_s$ (Empirical Fisher) | $\left(p_W(x) - e_y\right)\left(p_W(x) - e_y\right)^\top$ |
| $F_s$ (True Fisher) | $\mathrm{diag}(p_W(x)) - p_W(x)p_W(x)^\top$ |

Table 10: Output-space kernel $\mathcal{K}(x,y)$ inducing $C_{\text{grad}}$.

## D Computing the Curvature Matrix

The calibration set $\mathcal{X} = \{x_n\}_{n=1}^N$ consists of $N$ token sequences, with sequence $x_n$ of length $T$. The total number of token positions is $N_{\text{tok}} := NT$. For position $t$ in sequence $n$, the forward map decomposes as

$$x_n \longmapsto a_{n,t} \in \mathbb{R}^{n_a} \overset{\mathrm{L}_W}{\longmapsto} s_{n,t} \in \mathbb{R}^{n_s} \overset{q}{\longmapsto} f_{n,t} \in \mathbb{R}^V, \quad (38)$$

where $a_{n,t}$ is the input of the layer $\mathrm{L}_W : a \mapsto W^\top a$ at position $t$, $s_{n,t}$ is its output, and $q$ denotes the rest of the network up to the logits $f_{n,t}$. We denote by $h_{n,t} \in \mathbb{R}^D$ the input of the LM head, i.e. the last hidden state produced by $q$ before the linear projection to the vocabulary. The LM head is itself linear,

$$f_{n,t} = W_U h_{n,t}, \quad W_U \in \mathbb{R}^{V\times D}, \quad (39)$$

where $W_U$ denotes the LM head weight.

The matrix $\Sigma_{\text{grad}} \in \mathbb{R}^{n_s\times n_s}$ is defined as

$$\Sigma_{\text{grad}} := \mathbb{E}_{x\sim\mathcal{D}}\left(\left(\frac{\partial f_W}{\partial s}(x)\right)^\top\frac{\partial f_W}{\partial s}(x)\right). \quad (40)$$

For sequence inputs, the expectation in Equation 40 expands as a sum over all $(t,t')$ pairs with $t' \geq t$, the causal constraint $\partial f_{n,t'}/\partial s_{n,t} = 0$ for $t' < t$ removing the remainder:

$$\Sigma_{\text{grad}} = \frac{1}{N_{\text{tok}}}\sum_{n=1}^{N}\sum_{t=1}^{T}\sum_{t'=t}^{T}\left(J_{n,t,t'}\right)^\top J_{n,t,t'}, \quad (41)$$

$$\text{with} \quad J_{n,t,t'} := \frac{\partial f_{n,t'}}{\partial s_{n,t}} \in \mathbb{R}^{V\times n_s}.$$

Its direct evaluation is intractable: reverse-mode differentiation delivers one row of $J_{n,t,t'}$ per backward pass and isolates one output position $t'$ per cotangent assignment, so Equation 41 costs $T \times V$ backward passes per sequence, which is $6.6 \times 10^7$ for Llama 2-7B at $T = 2048$.

This cost is the product of two independent axes: the vocabulary index, of size $V$, and the output-position index, of size $T$. We remove the first exactly, down to the hidden size $D$, by absorbing the LM head into a Gram matrix (Appendix D.2); we then reduce both remaining axes to a single user-chosen budget $K$ by two applications of one randomization principle, stated next. The resulting estimator costs $K$ backward passes per sequence, independently of $V$ and $T$, and is unbiased for Equation 41.

### D.1 A Rademacher Annihilation Lemma

Both axes are removed by the same mechanism: a random signed superposition, whose second moment retains the diagonal terms of a sum of Gram matrices and annihilates the cross terms in expectation.

**Lemma 3.** *Let $A_1, \ldots, A_m \in \mathbb{R}^{p\times q}$ and let $\varepsilon_1, \ldots, \varepsilon_m$ be i.i.d. Rademacher variables, $\mathbb{P}[\varepsilon_i = \pm 1] = 1/2$. Then*

$$\mathbb{E}_\varepsilon\left[\left(\sum_{i=1}^m \varepsilon_i A_i\right)^\top\left(\sum_{i=1}^m \varepsilon_i A_i\right)\right] = \sum_{i=1}^m (A_i)^\top A_i. \quad (42)$$

*Proof.* Expanding the product gives $\sum_{i,i'} \varepsilon_i\varepsilon_{i'}(A_i)^\top A_{i'}$. Since the signs are independent and centred, $\mathbb{E}_\varepsilon[\varepsilon_i\varepsilon_{i'}] = \delta_{ii'}$, which annihi-lates every $i \neq i'$ term and preserves the $m$ diagonal terms. $\square$

The left-hand side of Equation 42 is computed from the single superposed matrix $\sum_i \varepsilon_i A_i$, whereas the right-hand side names $m$ separate matrices. Whenever the index $i$ is one that reverse-mode differentiation charges a backward pass for, Lemma 3 converts a cost of $m$ into a cost of one, at the price of a variance that we quantify in Appendix D.3. We apply it twice: with $i$ ranging over the vocabulary and with $i$ ranging over output positions (Appendix D.3).

### D.2 Eliminating the Vocabulary Dimension

The chain rule applied to the LM head $f_{n,t'} = W_U h_{n,t'}$ gives

$$J_{n,t,t'} = W_U H_{n,t,t'}, \tag{43}$$

with $H_{n,t,t'} := \frac{\partial h_{n,t'}}{\partial s_{n,t}} \in \mathbb{R}^{D \times n_s}$, so that $W_U$ factors out of every summand of Equation 41:

$$\Sigma_{\text{grad}} = \frac{1}{N_{\text{tok}}} \sum_{n,t} \sum_{t'=t}^{T} \left(H_{n,t,t'}\right)^\top M_G H_{n,t,t'}, \tag{44}$$

where $M_G := W_U^\top W_U \in \mathbb{R}^{D \times D}$. The vocabulary dimension is absorbed once and for all into the head Gram matrix $M_G$, which is precomputed at negligible cost.

The matrix $M_G$ is symmetric positive semi-definite, so it admits the spectral decomposition

$$M_G = \sum_{j=1}^{D} u_j \left(u_j\right)^\top, \quad u_j := \sqrt{\lambda_j} v_j, \tag{45}$$

where $\left(\lambda_j, v_j\right)_{j=1}^{D}$ are the eigenpairs of $M_G$. Substituting Equation 45 into Equation 44 gives

$$\begin{aligned}\Sigma_{\text{grad}} &= \frac{1}{N_{\text{tok}}} \sum_{n,t,t'} \sum_{j=1}^{D} \left(H_{n,t,t'}\right)^\top u_j \left(u_j\right)^\top H_{n,t,t'} \\ &= \frac{1}{N_{\text{tok}}} \sum_{n=1}^{N} \sum_{j=1}^{D} \sum_{t=1}^{T} \sum_{t'=t}^{T} \left(g_{n,t,t',j}\right)^\top g_{n,t,t',j},\end{aligned} \tag{46}$$

where

$$\begin{aligned} g_{n,t,t',j} &:= \left(u_j\right)^\top H_{n,t,t'} \qquad (47) \\ &= \frac{\partial\left(\left(u_j\right)^\top h_{n,t'}\right)}{\partial s_{n,t}} \in \mathbb{R}^{1 \times n_s}\end{aligned}$$

is the derivative at $s_{n,t}$ of the scalar $\left(u_j\right)^\top h_{n,t'}$, obtained by placing the cotangent $u_j$ at the LM-head input of the single output position $t'$.

The reduction is exact and lowers the cost per output position from $V$ to $D$ backward passes, a factor $V/D \approx 8$ for Llama 2-7B and $\approx 31$ for Llama 3-8B. The eigenbasis $\left(u_j\right)_{j=1}^{D}$ makes the $D$ required cotangent directions explicit, one backward pass each, and Equation 46 is the form from which the estimator of Appendix D.3 is built.

### D.3 The broadcast estimator

**Reducing the Backward Passes.** The factor $D$ is nonetheless the hidden size of the model, and is the dominant term in the calibration budget. We reduce it to a user-chosen $K \leq D$ by applying Lemma 3 along the vocabulary index. To this end we use, in place of the spectral decomposition Equation 45, the row decomposition of the same matrix $M_G = \sum_{v=1}^{V} w_v (w_v)^\top$ with $w_v^\top$ the $v$-th row of $W_U$. Draw a Rademacher matrix $\Psi \in \{\pm 1\}^{V \times K}$ and set $Z := W_U^\top \Psi \in \mathbb{R}^{D \times K}$, such that

$$z_k := W_U^\top \psi_k = \sum_{v=1}^{V} \Psi_{vk} w_v, \tag{48}$$

with $\psi_k$ the $k$-th column of $\Psi$. Lemma 3 applied with $A_v = (w_v)^\top$ yields, for each column independently,

$$\mathbb{E}_\Psi \left[z_k (z_k)^\top\right] = \sum_{v=1}^{V} w_v (w_v)^\top = M_G, \tag{49}$$

hence

$$\frac{1}{K} \mathbb{E}_\Psi [Z Z^\top] = M_G. \tag{50}$$

The $K$ columns of $Z$ replace the $D$ eigendirections $u_j$ of Equation 45 as cotangent directions.

**Eliminating the Output-Position Dimension.** Writing $g_{n,t,t',k} := (z_k)^\top H_{n,t,t'}$ for the projected gradient of Equation 47 along the direction $z_k$, a single backward pass differentiates a scalar that has already been summed over output positions: seeding the cotangent $z_k$ at every LM-head input $h_{n,t'}$ returns $\sum_{t' \geq t} g_{n,t,t',k}$ for all input positions $t$ at once. Squaring this quantity aggregates the displacements of distinct output positions before taking the norm, which is not the summand of Equation 46. Isolating one output position per backward pass restores it, at a cost of $T$ passes per direction.

We instead weight each output position by an independent random sign. Let $\varepsilon_{n,t'}^{(k)} \in \{\pm 1\}$ be i.i.d. Rademacher variables over $(n, t', k)$, drawn independently of $\Psi$, and consider the estimator

$$\tilde{\Sigma}_{\text{grad}} := \frac{1}{N_{\text{tok}}K}\sum_{n=1}^{N}\sum_{k=1}^{K}\sum_{t=1}^{T}\left(\tilde{g}_{n,t,k}\right)^{\top}\tilde{g}_{n,t,k}, \quad (51)$$

$$\text{with} \quad \tilde{g}_{n,t,k} := \sum_{t'=t}^{T}\varepsilon_{n,t'}^{(k)} g_{n,t,t',k} \in \mathbb{R}^{1\times n_s}. \quad (52)$$

The vector $\tilde{g}_{n,t,k}$ is the derivative at $s_{n,t}$ of the signed projected scalar $\sum_{t'}\varepsilon_{n,t'}^{(k)}(z_k)^{\top}h_{n,t'}$, so it is still obtained for all $t$ in one backward pass: the signs change the cotangent placed at each position, not the number of traversals of the computation graph. Per sequence the cost is one forward pass and $K$ backward passes, independent of both $V$ and $T$.

Applying Lemma 3 along the output-position index annihilates the cross-position terms $t' \neq t''$ in expectation and preserves the diagonal, so each summand of Equation 51 is unbiased for the corresponding summand of Equation 46 along the direction $z_k$.

**Proposition 4.** *Conditionally on the calibration set, $\mathbb{E}_{\Psi,\varepsilon}\left[\tilde{\Sigma}_{\text{grad}}\right] = \Sigma_{\text{grad}}$ entrywise, with $\Sigma_{\text{grad}}$ as in Equation 41. Moreover $\tilde{\Sigma}_{\text{grad}} \succeq 0$ in every realization of $(\Psi, \varepsilon)$.*

*Proof.* Fix $(n, t)$ and $k$. Conditioning on $\Psi$ and taking the expectation over $\varepsilon^{(k)}$ first, Lemma 3 with $A_{t'} = g_{n,t,t',k}$ gives $\mathbb{E}_{\varepsilon}\left[\left(\tilde{g}_{n,t,k}\right)^{\top}\tilde{g}_{n,t,k}\right] = \sum_{t'\geq t}\left(H_{n,t,t'}\right)^{\top} z_k z_k^{\top} H_{n,t,t'}$. Taking the expectation over $\Psi$ and applying Equation 50 replaces $z_k z_k^{\top}$ by $M_G$, which is Equation 44 term by term; the two randomizations commute because $\Psi$ and $\varepsilon$ are independent. Positive semidefiniteness is immediate from Equation 51, every summand being an outer product of a real row vector. □

The uniform seed does not admit the same statement. Writing $\hat{\Sigma}_{\text{grad}}$ for Equation 51 with $\varepsilon_{n,t'}^{(k)} \equiv 1$, and

$$S_{n,t',t''}^{(k)} := \sum_{t=1}^{T}\Big(\left(g_{n,t,t',k}\right)^{\top} g_{n,t,t'',k} + \left(g_{n,t,t'',k}\right)^{\top} g_{n,t,t',k}\Big) \in \mathbb{R}^{n_s\times n_s} \quad (53)$$

for the symmetrized contribution of the output-position pair $t' < t''$, the two estimators differ by a term that no choice of budget removes.

**Proposition 5.** *For every realization of $\Psi$,*

$$\hat{\Sigma}_{\text{grad}} = \mathbb{E}_{\varepsilon}\left[\tilde{\Sigma}_{\text{grad}}\right] + \frac{1}{N_{\text{tok}}K}\sum_{n=1}^{N}\sum_{k=1}^{K}\sum_{t'<t''} S_{n,t',t''}^{(k)}, \quad (54)$$

*and the second term is deterministic given the calibration set and the probes.*

| **Construction** | **Vocabulary axis** | **Backward passes** |
|---|---|---|
| Direct evaluation of Equation 41 | $V$ exact | $TV = 6.6 \times 10^7$ |
| Head-Gram spectral, Equation 46 | $D$ exact | $TD = 8.4 \times 10^6$ |
| $\tilde{\Sigma}_{\text{grad}}$, Equation 51 | $K$ probes | $K = 64$ |

Table 11: Calibration cost per sequence, in backward passes, for the constructions of Appendix D. Numerical column instantiated for Llama 2-7B ($V = 32000$, $D = 4096$, $T = 2048$) with $K = 64$.

*Proof.* Setting all signs to one in Equation 51 gives $\left(\tilde{g}_{n,t,k}\right)^{\top}\tilde{g}_{n,t,k} = \sum_{t',t''}\left(g_{n,t,t',k}\right)^{\top} g_{n,t,t'',k}$. Splitting the double sum into its diagonal $t' = t''$ and its off-diagonal part, and pairing $(t', t'')$ with $(t'', t')$ in the latter, gives Equation 54 with the diagonal part equal to $\mathbb{E}_{\varepsilon}\left[\tilde{\Sigma}_{\text{grad}}\right]$ by Lemma 3. No sign variable appears in the remainder. □

The second term of Equation 54 is an average over the $N$ calibration sequences of the cross-position blocks Equation 53. It does not decrease with the number of sequences or with the number of probes, and vanishes only if those blocks cancel on average.

As shown in Table 11, the proposed estimator $\tilde{\Sigma}_{\text{grad}}$ reduces the per-sequence calibration cost for Llama 2-7B from $TV \approx 6.6 \times 10^7$ backward passes for direct evaluation to $K = 64$, a reduction of about six orders of magnitude. On a single H200 GPU, the full **Complete-SVD** pipeline takes about 80 minutes for both Llama 3-8B and Mistral-7B (Table 12), most of which is spent estimating $\Sigma_{\text{grad}}$; since $\Sigma_{\text{input}}$ and $\Sigma_{\text{grad}}$ do not depend on the target rank, this cost is incurred once per model

| Model | $\Sigma_{\text{input}}$ | $\Sigma_{\text{grad}}$ | **SVD** | **Total** |
|---|---|---|---|---|
| Llama 3-8B | 12 min | 64 min | 5 min | ∼ 80 min |
| Mistral-7B | 12 min | 61 min | 5 min | ∼ 80 min |

Table 12: Per-stage wall-clock cost of the **Complete-SVD** compression pipeline on Llama 3-8B and Mistral-7B on a single H200 GPU: the input covariance estimation ($\Sigma_{\text{input}}$), the Gram matrix of the output Jacobians estimation ($\Sigma_{\text{grad}}$) and SVD computation. Reported with 256 calibration samples of length 2048.

| CR | $\Sigma_{\text{grad}}$ (exact) | $\hat{\Sigma}_{\text{grad}}$ | $\tilde{\Sigma}_{\text{grad}}$ |
|---|---|---|---|
| 0.90 | 23.4 | 23.8 | 23.6 |
| 0.85 | 25.6 | 26.2 | 25.9 |
| 0.80 | 29.0 | 29.8 | 29.6 |

Table 13: WikiText-2 perplexity of Pythia-410m compressed with the exact output-Jacobian Gram matrix $\Sigma_{\text{grad}}$ of Equation 41, with uniform seed estimator $\hat{\Sigma}_{\text{grad}}$, and with the proposed estimator $\tilde{\Sigma}_{\text{grad}}$ of Equation 51 computed each with 128 samples and 128 sequence length. The uncompressed model attains a perplexity of 15.83. Lower is better.

and calibration set and is shared across all compression ratios.

### D.4 Validation Against the Exact Reference

At reduced scale Equation 46 is directly computable, so the estimator can be validated end to end. Table 13 compares against this reference both $\tilde{\Sigma}_{\text{grad}}$ and the broadcast construction $\hat{\Sigma}_{\text{grad}}$, obtained from Equation 51 by replacing the random signs with $\varepsilon^{(k)}_{n,t'} \equiv 1$.

As shown in Table 13, the exact reference is the best perplexity attainable under this weighting and both estimators sit above it. Relative to $\hat{\Sigma}_{\text{grad}}$, the proposed estimator $\tilde{\Sigma}_{\text{grad}}$ closes the gap to the reference at each compression ratio.

### D.5 Extension to a Tractable Fisher Matrix Estimator

The construction of Appendix D extends to the true Fisher matrix of the model, obtained by replacing the identity weighting of Equation 41 with the curvature of the predictive distribution. Write $p_{n,t'} := \mathrm{Softmax}(f_{n,t'}) \in \mathbb{R}^V$ for the predictive distribution at output position $t'$ of sequence $n$, and

$$C_{n,t'} := \mathrm{diag}(p_{n,t'}) - p_{n,t'}(p_{n,t'})^\top \in \mathbb{R}^{V\times V} \tag{55}$$

for the Hessian of the negative log-likelihood with respect to the logits, evaluated at $f_{n,t'}$. Taking the contexts from the calibration set and the labels from the model itself, the score of the sequence log-likelihood decomposes over output positions, and the resulting Fisher matrix at the layer pre-activation is

$$F_s = \frac{1}{N_{\text{tok}}} \sum_{n=1}^{N} \sum_{t=1}^{T} \sum_{t'=t}^{T} (J_{n,t,t'})^\top C_{n,t'} J_{n,t,t'}, \tag{56}$$

with $J_{n,t,t'}$ as in Equation 41. No cross-output-position terms appear: conditionally on the context, the sampled labels at distinct positions are independent and each per-position score is centred, so the corresponding expectations vanish. Equation 56 is therefore exact, and plays for the Fisher matrix the role that Equation 41 plays for $\Sigma_{\text{grad}}$.

The vocabulary-elimination step of Appendix D.2 does not transfer. Substituting Equation 43 into Equation 56 gives $(H_{n,t,t'})^\top M_{n,t'} H_{n,t,t'}$ with $M_{n,t'} := W_U^\top C_{n,t'} W_U \in \mathbb{R}^{D\times D}$, but $M_{n,t'}$ now depends on the output position, so it cannot be factored out of the sum as the constant $M_G$ was. Recovering the eigenbasis construction of Equation 46 would require one spectral decomposition and $D$ cotangent directions per output position, hence $TD$ backward passes per sequence, which is the cost Appendix D.3 was designed to remove.

**Probes matched to the predictive distribution.** We instead randomize along the vocabulary index as in Equation 48, but with a probe distribution matched to $C_{n,t'}$ rather than to $M_G$. Let $e_v \in \mathbb{R}^V$ denote the $v$-th canonical basis vector. The matrix Equation 55 admits the exact expansion

$$\begin{aligned} C_{n,t'} &= \sum_{v=1}^{V} p_{n,t',v}(e_v - p_{n,t'})(e_v - p_{n,t'})^\top \\ &= \mathbb{E}_{y\sim p_{n,t'}}\left[(e_y - p_{n,t'})(e_y - p_{n,t'})^\top\right], \end{aligned} \tag{57}$$

where $p_{n,t',v}$ is the $v$-th entry of $p_{n,t'}$ and the expectation is over a single categorical draw $y$ from $p_{n,t'}$. The weighted sum over the vocabulary is thus a plain expectation over a term visited at its own weight, and the budget $K$ now indexes independent draws rather than Rademacher columns. Let $y^{(k)}_{n,t'} \sim p_{n,t'}$ be independent over $(n, t', k)$, and define the cotangent directions

$$z^{(k)}_{n,t'} := W_U^\top \left(e_{y^{(k)}_{n,t'}} - p_{n,t'}\right) \tag{58}$$

$$= w_{y^{(k)}_{n,t'}} - W_U^\top p_{n,t'} \in \mathbb{R}^D, \tag{59}$$

with $w_v$ the $v$-th row of $W_U$ as in Appendix D.3. Each direction is one row of the LM head corrected by the single vector $W_U^\top p_{n,t'}$, so beyond the logits themselves no object of size $V \times V$ or $V \times D$ is formed at any position, and Equation 57 gives $\mathbb{E}_y\left[z^{(k)}_{n,t'}\left(z^{(k)}_{n,t'}\right)^\top\right] = M_{n,t'}$.

**The estimator.** Unlike the probes Equation 48, the directions Equation 58 vary with the output position, which is what allows the position-dependent curvature to be represented. Writing $\gamma_{n,t,t',k} := \left(z^{(k)}_{n,t'}\right)^{\top} H_{n,t,t'}$ for the projected gradient of Equation 47 along $z^{(k)}_{n,t'}$, the estimator is

$$\tilde{F}_s := \frac{1}{N_{\text{tok}} K} \sum_{n=1}^{N} \sum_{k=1}^{K} \sum_{t=1}^{T} \left(\tilde{\gamma}_{n,t,k}\right)^{\top} \tilde{\gamma}_{n,t,k} \quad (60)$$

$$\text{with} \quad \tilde{\gamma}_{n,t,k} := \sum_{t'=t}^{T} \gamma_{n,t,t',k} \in \mathbb{R}^{1 \times n_s}. \quad (61)$$

As in Equation 51, $\tilde{\gamma}_{n,t,k}$ is obtained for all input positions $t$ in a single backward pass, by seeding the cotangent $z^{(k)}_{n,t'}$ at the LM-head input of every output position $t'$. The probes are held fixed in this differentiation: although $z^{(k)}_{n,t'}$ is a function of $p_{n,t'}$ and hence of $h_{n,t'}$, the scalar differentiated is $\sum_{t'=1}^{T} \left(z^{(k)}_{n,t'}\right)^{\top} h_{n,t'}$ with each $z^{(k)}_{n,t'}$ treated as a constant, so that Equation 58 enters through the cotangent alone and not through the computation graph.

No sign variables appear in Equation 61. The role played by Lemma 3 along the output-position index in Equation 51 is here played by the probe construction itself: the directions Equation 58 are drawn independently at each output position and satisfy $\mathbb{E}_y\left[z^{(k)}_{n,t'}\right] = 0$, so the cross-position terms are annihilated in expectation without an auxiliary randomization. The estimator Equation 51 requires the signs for a different reason. Its probe $z_k$ is centred as well, but the same realization is seeded at every output position, so the relevant moment for $t' \neq t''$ is $\mathbb{E}_{\Psi}\left[z_k z_k^{\top}\right] = M_G \neq 0$ and the cross terms survive; independent signs restore the cancellation that independent draws provide here for free.

**Proposition 6.** *Conditionally on the calibration set, $\mathbb{E}_y\left[\tilde{F}_s\right] = F_s$ entrywise, with $F_s$ as in Equation 56. Moreover $\tilde{F}_s \succeq 0$ in every realization of the draws.*

*Proof.* Fix $(n, t)$ and $k$. Expanding the outer product in Equation 61 over two output-position indices gives $\sum_{t' \geq t} \sum_{t'' \geq t} \left(H_{n,t,t'}\right)^{\top} z^{(k)}_{n,t'} \left(z^{(k)}_{n,t''}\right)^{\top} H_{n,t,t''}$. The draws being independent across output positions and centred, $\mathbb{E}_y\left[z^{(k)}_{n,t'} \left(z^{(k)}_{n,t''}\right)^{\top}\right]$ equals $M_{n,t'}$ when $t' = t''$ and vanishes otherwise, leaving $\sum_{t' \geq t} \left(H_{n,t,t'}\right)^{\top} M_{n,t'} H_{n,t,t'}$, which is Equation 56 term by term by Equation 43. Positive semidefiniteness is immediate from Equation 61, every summand being an outer product of a real row vector. □

The cost per sequence is one forward pass and $K$ backward passes, independent of $V$ and $T$, matching Table 11. The forward pass supplies both the predictive distributions entering Equation 58 and the graph traversed by the backward passes, and no spectral decomposition is required at any stage. The budget $K$ controls the variance alone: Proposition 6 holds for every $K \geq 1$, so the centre is fixed and the variance decreases as $O(1/K)$, in contrast to constructions that restrict Equation 57 to a fixed subset of the vocabulary and renormalize $p_{n,t'}$ on it, whose error is deterministic and does not vanish with the calibration budget.

### D.5.1 Comparison with the Top-$m$ Construction of IO-SVD

IO-SVD targets the same output-side curvature as Equation 56 and, like the present work, pairs it with an input activation covariance under a Kronecker factorization. It differs in how the two intractable axes are removed, and both of its reductions introduce a deterministic deviation from Equation 56 rather than a variance that decreases with the calibration budget.

**The output-position axis.** IO-SVD defines its output-side curvature through the same-position Jacobian $\partial z_t / \partial h_{\ell,t}$, so its accumulated object is the single term $t' = t$ of Equation 56:

$$\frac{1}{N_{\text{tok}}} \sum_{n=1}^{N} \sum_{t=1}^{T} \left(J_{n,t,t}\right)^{\top} C_{n,t} J_{n,t,t}. \quad (62)$$

In a causal model the pre-activation $s_{n,t}$ influences the logits at every position $t' \geq t$, so a perturbation at input position $t$ contributes to $T - t + 1$ terms of Equation 56 and Equation 62 retains one of them. The omission is structural. It does not depend on the vocabulary budget $m$, it is not reduced by enlarging the calibration set, and its extent grows with sequence length.

**The vocabulary axis.** IO-SVD restricts Equation 55 to the $m$ largest entries of the predictive distribution and renormalizes the probabilities on that support. Fix a sequence and an output position and suppress the indices $(n, t')$. Let $S$ collect the $m$ largest entries of $p$, let $p_S \in \mathbb{R}^m$ be the restriction of $p$ to $S$, let $\pi := \mathbf{1}^{\top} p_S$ be its mass, and let $C_S$ be the principal submatrix of $C$ indexed by $S$. The renormalized curvature is $C^{\text{IO}} := \text{diag}(p_S/\pi) - p_S p_S^{\top}/\pi^2$, and a direct computation gives

| CR | Method | Mistral-7B | | | | Llama 3-8B | | | | Qwen3-8B | | | |
|---|---|---|---|---|---|---|---|---|---|---|---|---|---|
| | | Wiki ↓ | C4 ↓ | PTB ↓ | Acc ↑ | Wiki ↓ | C4 ↓ | PTB ↓ | Acc ↑ | Wiki ↓ | C4 ↓ | PTB ↓ | Acc ↑ |
| | Baseline | 5.3 | 8.1 | 27.7 | 64.0 | 6.1 | 9.5 | 9.9 | 63.3 | 9.7 | 15.5 | 15.4 | 62.0 |
| 0.9 | KFAC-SVD | 6.2 | 13.1 | 44.0 | 56.9 | 8.8 | 25.7 | 32.6 | 54.5 | 11.5 | 23.8 | 28.0 | 57.1 |
| | IO-SVD | **6.1** | 12.8 | 40.0 | 56.9 | 8.4 | 23.2 | 30.9 | 56.2 | **11.0** | 23.7 | 28.9 | **57.9** |
| | **KL-SVD** | **6.1** | 12.8 | 40.1 | **57.3** | **8.3** | 22.8 | 27.1 | **56.9** | 11.2 | 24.6 | 29.1 | 57.7 |
| | **Complete-SVD** | 6.4 | **12.7** | **39.9** | 56.9 | 8.7 | **22.4** | **23.0** | 55.8 | 11.7 | **23.4** | **23.4** | 56.7 |
| 0.7 | KFAC-SVD | 9.3 | 34.6 | 110.7 | 45.8 | 19.4 | 151.3 | 253.5 | 40.7 | 14.7 | 50.0 | 94.0 | 46.8 |
| | IO-SVD | **8.8** | 33.1 | 97.8 | 45.6 | 16.8 | 111.8 | 219.2 | 43.3 | 14.4 | 58.7 | 120.2 | **49.2** |
| | **KL-SVD** | 10.3 | 34.7 | 98.8 | **46.6** | **16.0** | 103.0 | 172.7 | **45.6** | **14.2** | 49.9 | 82.8 | 48.6 |
| | **Complete-SVD** | 10.1 | **32.9** | **90.7** | 45.4 | 16.8 | **95.3** | **123.7** | 44.2 | 15.0 | **47.5** | **67.7** | 47.8 |

Table 14: Output-aware methods from Table 1 (WikiText-2 calibration, uniform rank allocation), repeated here for the comparison of **KL-SVD** with IO-SVD. Accuracy (Acc) is the average zero-shot accuracy over 6 downstream datasets.

$$C^{\mathrm{IO}} = \frac{1}{\pi}\Big(C_S - \frac{1-\pi}{\pi} p_S p_S^\top\Big). \tag{63}$$

The truncation therefore does more than set the entries outside $S$ to zero: it inflates the retained block by $\pi^{-1}$ and over-subtracts the rank-one term by $\pi^{-2}$. Since $\pi$ varies from one output position to the next, Equation 63 reweights positions against one another rather than applying a common scale, and the traces differ, $\mathrm{Tr}(C^{\mathrm{IO}}) = 1 - \|p_S/\pi\|^2$ against $\mathrm{Tr}(C) = 1 - \|p\|^2$. Restricting Equation 57 to $S$ without renormalizing would instead give the partial sum $\sum_{v\in S} p_v(e_v - p)(e_v - p)^\top$, which is the head term of a stratified form of our estimator: positive semi-definite, free of the distortion Equation 63, and monotone in $m$.

Both deviations are deterministic given the calibration set. They persist at any $m$ short of $V$ and at any number of calibration sequences, so the corresponding error is a bias rather than a variance. By contrast, Proposition 6 places the centre of $\tilde{F}_s$ at Equation 56 for every $K \geq 1$, leaving only an $O(1/K)$ variance, and the two constructions coincide neither in the output positions they retain nor in the curvature they assign to those retained.

### D.5.2 SVD Compression with KL Divergence Objective

To further validate the design of **Complete-SVD**, we compare it against an alternative objective based on the Kullback-Leibler (KL) divergence between the original and compressed model output distributions discussed in Appendix B.4. The KL-based objective is connected to the true Fisher information matrix, which arises as the second-order expansion of the KL divergence and constitutes the principled curvature metric for probabilistic models. Furthermore, in our experiment, we use the unbiased estimator of True Fisher information matrix described in Appendix D.5.

The configurations in Table 14 share the rank allocation and covariance regularization procedure (Appendix F). Compared with IO-SVD, **KL-SVD** yields gains that depend on the model and compression ratio. On Llama 3-8B, it lowers perplexity on all three corpora at both ratios, with larger improvements under stronger compression: at ratio 0.7, C4 perplexity decreases from 111.8 to 103.0 and PTB perplexity from 219.2 to 172.7, corresponding to reductions of 7.9% and 21.2%. On Qwen3-8B, IO-SVD is marginally better on all three corpora at ratio 0.9, whereas **KL-SVD** improves all three perplexities at ratio 0.7, reducing C4 and PTB perplexity by 15.0% and 31.1%, respectively. On Mistral-7B, the two methods are nearly identical in perplexity at ratio 0.9, whereas **KL-SVD** yields higher perplexity on all three corpora at ratio 0.7, particularly on WikiText-2 (10.3 versus 8.8). Average zero-shot accuracy is higher with **KL-SVD** on Mistral-7B and Llama 3-8B at both ratios, but slightly lower on Qwen3-8B. Furthermore, on IFEval (Table 2), **KL-SVD** improves prompt-level strict accuracy over IO-SVD by 2.8– 8.2 percentage points in all but one setting; the exception is Mistral-7B at ratio 0.9, where accuracy decreases from 13.9 to 12.4.

## E Remapping storage

Standard SVD compression stores the truncated factors $\tilde{U}_k \in \mathbb{R}^{n_s \times k}$ and $\tilde{V}_k \in \mathbb{R}^{n_a \times k}$ separately, with footprint $k(n_s + n_a)$ and parameter ratio $\alpha = k(n_s + n_a)/n_s n_a$. This ratio saturates before $k$ reaches $\min(n_s, n_a)$ (for $n_s = n_a$, $\alpha = 1$ already at $k = n_a/2$), so $\alpha$ is not in one-to-one correspondence with $k$ over $(0, 1]$. Following Dobi-SVD (Wang et al., 2025b), assuming $n_s \geq n_a$, we adopt a remapping scheme that halves the precision of the overlapping $n_a \times k$ submatrices of $\tilde{U}_k$ and $\tilde{V}_k$ and merges them into a single $n_s \times k$ matrix, yielding a parameter ratio $\tilde{\alpha} = k/\min(n_s, n_a)$ in one-to-one correspondence with $k$. Under remapping, we thus choose the ranks uniformly as $k = \lfloor \tilde{\alpha} \cdot \min(n_s, n_a) \rfloor$.

# F Implementation Details: Covariance Square Roots

**Regularization following KFAC-SVD.** We follow the covariance regularization and Cholesky factorization used in KFAC-SVD for both the input covariance and the output Jacobian-Gram covariance. For a covariance matrix $M \in \mathbb{R}^{d \times d}$, we first apply $M \leftarrow (1-\alpha)M + \alpha\left(\frac{\mathrm{tr}(M)}{d}\right)I$, using $\alpha = 0.1$ for the input covariance and $\alpha = 0.3$ for $\Sigma_{\mathrm{grad}}$. We then compute its Cholesky factorization $M = LL^\top$. If the Cholesky decomposition fails because of finite-precision negative eigenvalues, we compute the smallest eigenvalue $s_{\min}$ and shift the diagonal as $M \leftarrow M + (-s_{\min} + 10^{-6})I$ before retrying the factorization.

# G Inference Efficiency

**Inference throughput.** We measure the generation throughput, in tokens per second, of Llama 2-7B (Base) compressed by **Complete-SVD** at ratios of 40%, 60%, and 80% (Figure 3). On an NVIDIA A100 (80 GB), compression increases throughput at every ratio, with the largest relative gains at large batch sizes and short sequence lengths, consistent with the reduced parameter footprint of the linear layers. On a memory-limited NVIDIA Quadro P5000 (16 GB), which can still hold the original model, the model at 40% ratio is about 4 × faster than the original at batch size 128. At batch size 256, the original model runs out of memory, whereas the models at 40% and 60% ratios still run and reach their highest throughput.

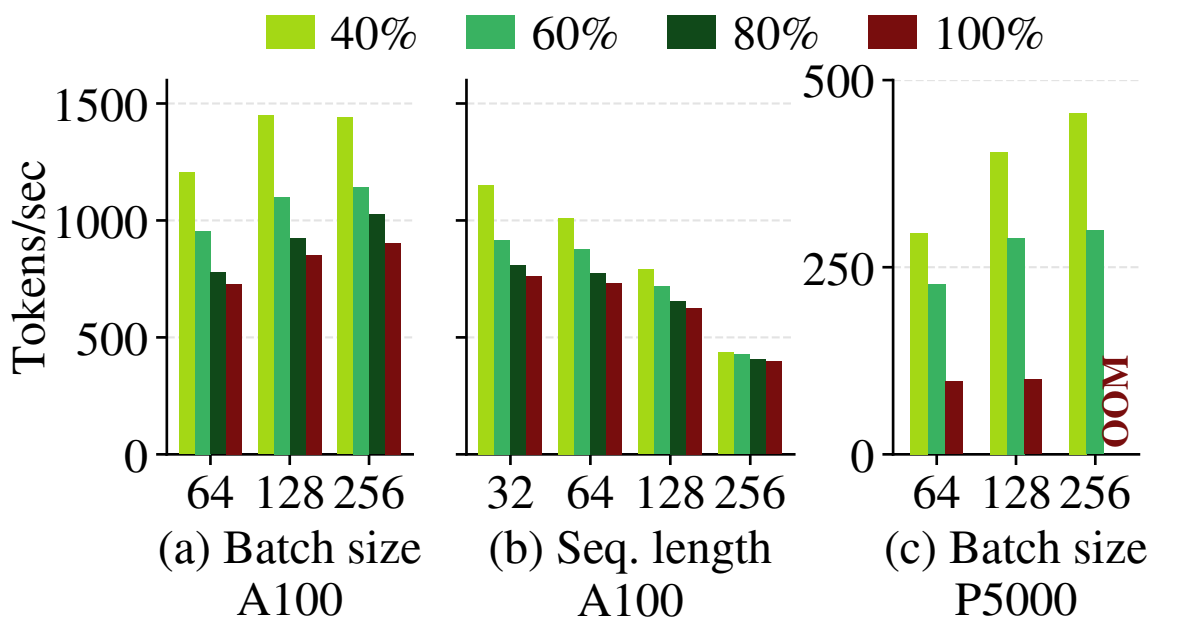

Figure 3: Inference throughput (tokens/sec) of Llama 2-7B (100%) and its **Complete-SVD** compressed variants at 40%, 60%, and 80% ratios. **(a)** Varying batch size with sequence length 32 on an A100 (80GB). **(b)** Varying sequence length with batch size 64 on an A100. **(c)** Varying batch size with sequence length 32 on a Quadro P5000 (16GB). "OOM" denotes an out-of-memory failure of the baseline.

# H Experiments on VLMs

**ScienceQA-IMG** is the image-based subset of the ScienceQA test split. It contains 2,017 multiple-choice science questions from elementary and high-school curricula, covering natural, social and language science; each question comes with an image and two to five answer options, and performance is reported as top-1 accuracy (%) of the selected option.

**SEED-Bench** is a human-annotated multiple-choice benchmark for multimodal comprehension. We use its image subset (SEED-Bench-IMG), whose four-option questions span nine dimensions: scene understanding, instance identity, instance attributes, instance location, instance counting, spatial relation, instance interaction, visual reasoning and text recognition; performance is reported as top-1 accuracy (%).

**Complete-SVD$^{\mathrm{Q}}$.** **Complete-SVD** allocates ranks uniformly, so that every compressed projection retains the same fraction of its parameters. **Complete-SVD$^{\mathrm{Q}}$** keeps the **Complete-SVD** factorization unchanged and replaces this uniform allocation with the sensitivity-aware rank allocator of QSVD, which separates the contribution of the factorization from that of the rank allocation. For a projection $W \in \mathbb{R}^{m \times n}$, let $\Sigma_{\mathrm{grad}}^{\frac{1}{2}} W \Sigma_{\mathrm{input}}^{\frac{1}{2}} = \sum_i \sigma_i \boldsymbol{u}_i \boldsymbol{v}_i^\top$ denote the singular value decomposition of the whitened weight, so that $W = \sum_i \sigma_i \Sigma_{\mathrm{grad}}^{-\frac{1}{2}} \boldsymbol{u}_i \boldsymbol{v}_i^\top \Sigma_{\mathrm{input}}^{-\frac{1}{2}}$. Each component is scored by $I_i = \sigma_i \left( \frac{1}{N} \sum_{n=1}^{N} g_{n,i}^2 \right)^{\frac{1}{2}}$, where

$$g_{n,i} = \partial \mathcal{L}_n / \partial \sigma_i = \langle \nabla_W \mathcal{L}_n, \Sigma_{\text{grad}}^{-\frac{1}{2}} \boldsymbol{u}_i \boldsymbol{v}_i^\top \Sigma_{\text{input}}^{-\frac{1}{2}} \rangle \tag{64}$$

is the derivative of the cross-entropy of the $n$-th calibration sample with respect to $\sigma_i$, obtained with the gradient pass of QSVD on the same $N = 256$ calibration samples used to estimate $\Sigma_{\text{input}}$ and $\Sigma_{\text{grad}}$. As in QSVD, the components of all query, key and value projections are ranked jointly and the $k$ highest-scoring ones are retained, regardless of their position in the spectrum. Following the parameter-matched budget of WSVD, $k = \left\lfloor \rho \sum_\ell m_\ell n_\ell / (m_\ell + n_\ell) \right\rfloor$, so that the factorized projections hold a fraction $\rho$ of the original parameters. Let $\mathcal{I} \subset \{1, ..., \min(m, n)\}$ denote the indices of the components retained for a projection, which need not be the leading ones, and let $U_\mathcal{I}$, $V_\mathcal{I}$ and $\boldsymbol{\sigma}_\mathcal{I}$ collect the corresponding singular vectors and values. The factors are then $A = \text{diag}\,(\boldsymbol{\sigma}_\mathcal{I})^{\frac{1}{2}} V_\mathcal{I}^\top \Sigma_{\text{input}}^{-\frac{1}{2}}$ and $B = \Sigma_{\text{grad}}^{-\frac{1}{2}} U_\mathcal{I}\, \text{diag}\,(\boldsymbol{\sigma}_\mathcal{I})^{\frac{1}{2}}$, with rank $|\mathcal{I}|$; uniform **Complete-SVD** corresponds to $\mathcal{I} = \{1, ..., r\}$.

Table 15: Accuracy evaluation of different methods under FP16.

| Model | Method | ScienceQA-IMG ↑ | | | | | SEED-Bench ↑ | | | | | Avg. ↑ |
|---|---|---|---|---|---|---|---|---|---|---|---|---|
| | | 90% | 80% | 70% | 60% | 50% | 90% | 80% | 70% | 60% | 50% | |
| LLaVA1.5 7B | ASVD | 49.93 | 50.12 | 47.10 | 36.69 | 19.19 | 54.27 | 53.53 | 48.35 | 37.17 | 24.17 | 42.05 |
| | SVD-LLM | 65.44 | 63.71 | 61.92 | 57.41 | 55.53 | 57.89 | 57.50 | 55.33 | 54.64 | 55.31 | 58.47 |
| | QSVD | 67.72 | 68.22 | 67.08 | 65.05 | 62.37 | 59.84 | 59.07 | 59.78 | 59.00 | 58.23 | 62.64 |
| | WSVD | 68.17 | 67.72 | 67.28 | 65.89 | 65.49 | 60.10 | 60.17 | 59.89 | 60.18 | 60.46 | 63.54 |
| | IO-SVD | 67.92 | 68.07 | 68.17 | 66.73 | 65.99 | 60.29 | 60.32 | 60.22 | 60.15 | 60.64 | 63.85 |
| | **Complete-SVD** | 66.48 | 66.58 | 66.83 | 65.79 | 66.39 | 58.98 | 58.51 | 58.09 | 56.19 | 56.72 | 62.06 |
| | **Complete-SVD$^{Q}$** | 67.28 | 66.68 | 66.44 | 66.14 | 65.34 | 60.05 | 59.87 | 59.84 | 59.42 | 59.21 | 63.03 |
| | FP16 | Accuracy: 68.01 | | | | | Accuracy: 60.18 | | | | | 64.10 |
| LLaVA1.5 13B | ASVD | 71.39 | 71.59 | 70.00 | 70.25 | 69.51 | 61.92 | 61.91 | 61.54 | 61.51 | 60.71 | 66.03 |
| | SVD-LLM | 71.05 | 70.85 | 70.30 | 70.35 | 70.30 | 62.28 | 62.34 | 62.25 | 62.08 | 63.01 | 66.48 |
| | QSVD | 71.89 | 71.99 | 71.49 | 71.54 | 71.39 | 62.61 | 62.64 | 62.82 | 62.63 | 62.52 | 67.15 |
| | WSVD | 71.99 | 71.84 | 72.53 | 71.59 | 71.44 | 62.52 | 62.68 | 62.38 | 62.37 | 62.37 | 67.17 |
| | IO-SVD | 71.79 | 72.14 | 72.29 | 72.14 | 72.68 | 62.64 | 62.49 | 62.46 | 62.45 | 62.29 | 67.34 |
| | **Complete-SVD** | 71.15 | 71.99 | 71.89 | 72.09 | 71.59 | 62.58 | 62.65 | 62.44 | 62.54 | 62.30 | 67.12 |
| | **Complete-SVD$^{Q}$** | 71.49 | 71.59 | 71.79 | 71.94 | 71.44 | 62.41 | 62.53 | 62.44 | 62.44 | 62.54 | 67.06 |
| | FP16 | Accuracy: 71.83 | | | | | Accuracy: 62.53 | | | | | 67.18 |
| SmolVLM 2B | | 90% | 80% | 70% | | | 90% | 80% | 70% | | | |
| | ASVD | 29.30 | 3.97 | 0.20 | | | 17.85 | 1.50 | 0.95 | | | 8.96 |
| | SVD-LLM | 40.06 | 17.20 | 3.82 | | | 32.49 | 15.89 | 4.60 | | | 19.01 |
| | QSVD | 77.00 | 62.77 | 42.59 | | | 64.80 | 50.46 | 36.24 | | | 55.64 |
| | WSVD | 76.30 | 71.74 | 60.93 | | | 65.78 | 63.29 | 54.45 | | | 65.42 |
| | IO-SVD | **83.34** | 82.65 | 82.30 | | | 68.36 | **68.65** | **68.15** | | | 75.58 |
| | **Complete-SVD** | 79.33 | 76.55 | 73.77 | | | 66.67 | 65.69 | 64.63 | | | 71.11 |
| | **Complete-SVD$^{Q}$** | 83.14 | **83.59** | **82.55** | | | **68.64** | 68.39 | 68.02 | | | **75.72** |
| | FP16 | Accuracy: 84.53 | | | | | Accuracy: 68.47 | | | | | 76.53 |